%% file: main.tex
\documentclass{article}

\usepackage{microtype}
\usepackage{graphicx}
\usepackage{subcaption}
\usepackage{booktabs} 

\usepackage{hyperref}

\usepackage[accepted]{icml2026}

\usepackage{amsmath}
\usepackage{amssymb}
\usepackage{mathtools}
\usepackage{amsthm}
\usepackage{pifont}
\input{math}

\usepackage{multirow}
\usepackage{arydshln}
\usepackage[table]{xcolor}

\usepackage[capitalize,noabbrev]{cleveref}

\theoremstyle{plain}
\newtheorem{theorem}{Theorem}[section]

\newtheorem{lemma}[theorem]{Lemma}

\theoremstyle{definition}

\theoremstyle{remark}

\usepackage[textsize=tiny]{todonotes}

\icmltitlerunning{Dissecting Quantization Error: A Concentration-Alignment Perspective}

\begin{document}
\title{}
\twocolumn[
  \icmltitle{Dissecting Quantization Error: A Concentration-Alignment Perspective}



  \icmlsetsymbol{equal}{*}

  \begin{icmlauthorlist}
    \icmlauthor{Marco Federici}{comp}
    \icmlauthor{Boris van Breugel}{comp}
    \icmlauthor{Paul Whatmough}{comp}
    \icmlauthor{Markus Nagel}{comp}
  \end{icmlauthorlist}

  \icmlaffiliation{comp}{Qualcomm AI Research. Qualcomm AI Research is an initiative of Qualcomm Technologies, Inc}

  \icmlcorrespondingauthor{}{\{mfederic,bvanbreu,pwhatmou,markusn\}@qualcomm.com}

  \icmlkeywords{Machine Learning, ICML}

  \vskip 0.3in
]

\newcommand{\cmark}{\textcolor{green!60!black}{\ding{51}}} 
\newcommand{\xmark}{\textcolor{red!70!black}{\ding{55}}}   



\printAffiliationsAndNotice{}  

\begin{abstract}
  Quantization can drastically increase the efficiency of large language and vision models, but typically incurs an accuracy drop. Recently, function-preserving transforms (e.g. rotations, Hadamard transform, channel-wise scaling) have been successfully applied to reduce post-training quantization error, yet a principled explanation remains elusive.
  We analyze linear-layer quantization via the signal-to-quantization-noise ratio (SQNR), showing that for uniform integer quantization at a fixed bit width, SQNR decomposes into (i) the concentration of weights and activations (capturing spread and outliers), and (ii) the alignment of their dominant variation directions. This reveals an actionable insight: beyond concentration --- the focus of most prior transforms (e.g. rotations or Hadamard) --- improving alignment between weight and activation can further reduce quantization error. Motivated by this, we introduce block Concentration–Alignment Transforms (CAT), a lightweight linear transformation that uses a covariance estimate from a small calibration set to jointly improve concentration and alignment, approximately maximizing SQNR. Experiments across several LLMs show that CAT consistently matches or outperforms prior transform-based quantization methods at 4-bit precision, confirming the insights gained in our framework.
 
\end{abstract}

\input{sections/introduction}

\input{sections/method}
\input{sections/related_work}
\input{sections/experiments}
\input{sections/conclusions}




\section*{Impact Statement}

This paper's goal is to improve the understanding of quantization error, and provide tools for creating more efficient AI models. This reduces the environmental footprint, latency, and economical cost of AI, and may improve access to AI in resource-constraint environments. However, it is important to recognize that quantization may affect biases of LLMs.


\bibliography{bibliography}
\bibliographystyle{icml2026}

\newpage
\appendix
\onecolumn

\input{appendix/proofs}
\input{appendix/additional_results}


\end{document}

%% file: math.tex
\usepackage{amsmath,amsfonts,bm}

\newcommand{\E}[1]{\mathbb{E}\left[#1\right]}
\newcommand{\defined}{\stackrel{def}{=}}
\newcommand{\range}[1]{r\!\left(#1\right)}
\newcommand{\SQNR}[1]{\text{SQNR}\!\left(#1\right)}
\newcommand{\quant}[1]{\widetilde{#1}}
\newcommand{\diag}[1]{\text{Diag}\left(#1\right)}

\newcommand{\covariance}{{\boldsymbol{\Sigma}}_x}
\newcommand{\wcovariance}{{\boldsymbol{\Sigma}}_w}
\newcommand{\ycovariance}{{\boldsymbol{\Sigma}}_y}
\newcommand{\trace}[1]{\text{Tr}\left(#1\right)}
\newcommand{\eye}{\mI}
\newcommand{\uniform}[1]{\text{U}\left(#1\right)}
\newcommand{\discrete}[2]{\text{Discrete}\left(#1,#2\right)}

\def\eqref#1{equation~\ref{#1}}

\def\1{\bm{1}}

\def\rw{{\textnormal{w}}}

\def\vm{{\bm{m}}}

\def\vw{{\bm{w}}}
\def\vx{{\bm{x}}}

\def\mA{{\bm{A}}}
\def\mB{{\bm{B}}}

\def\mH{{\bm{H}}}
\def\mI{{\bm{I}}}

\def\mM{{\bm{M}}}

\def\mR{{\bm{R}}}

\def\mT{{\bm{T}}}

\def\mW{{\bm{W}}}

\DeclareMathAlphabet{\mathsfit}{\encodingdefault}{\sfdefault}{m}{sl}
\SetMathAlphabet{\mathsfit}{bold}{\encodingdefault}{\sfdefault}{bx}{n}

\DeclareMathOperator*{\argmax}{arg\,max}

%% file: sections/introduction.tex
\section{Introduction}

\begin{figure}[!tb]
    \centering
    \hspace*{-5mm}
    \includegraphics[width=1.05\linewidth, trim={3cm 0 0 0}, clip]{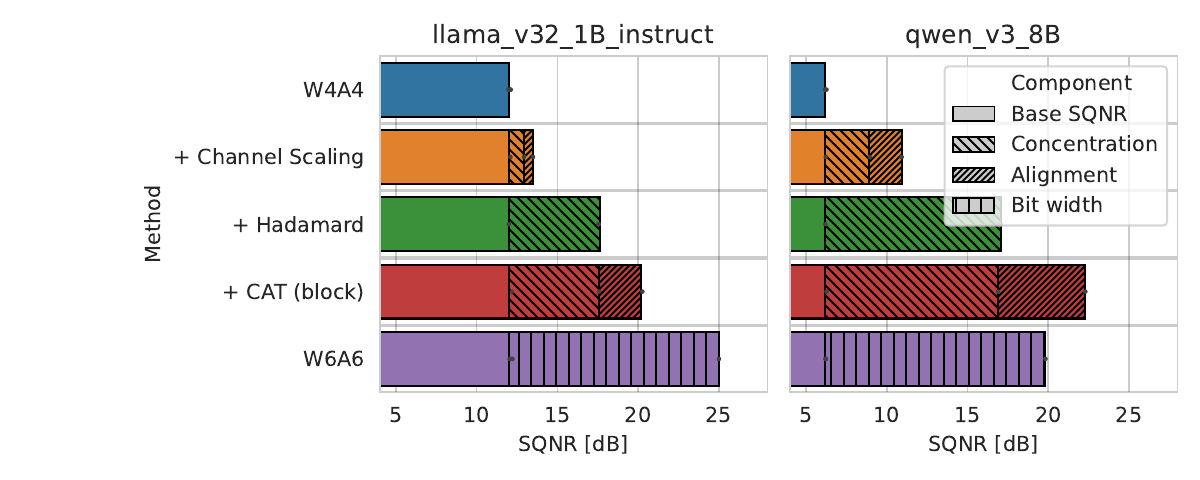}
    \caption{The signal-to-quantization-noise (SQNR) of a quantized linear layer can be factorized into a bit width term, concentration, and alignment. Orthogonal transforms (e.g. Hadamard, rotation) can improve the concentration (reduce outliers), but not the alignment. Concentration-Alignment Transform (\textbf{CAT}) is designed to improve alignment too, which yields a W4A4 SQNR that often rivals W6A6 quantization. }
    \label{fig:figone}
\end{figure}

Model quantization is the most fundamental strategy to reduce computation and memory requirements for large models such as LLMs. However, drastic reduction in performance is often observed when quantizing model weights and activations at lower bit-widths. Recently, the introduction of invertible linear transformations~\citep{xiao_smoothquant_2024,ashkboos_quarot_2024,van2025fptquant}, which are partially fused into the model weights, have enabled more aggressive compression at the small cost of additional operations during inference.
Popular linear transforms include rotations \citep{ashkboos_quarot_2024, liu_spinquant_2024, tseng_quip_2024}, element-wise scaling \citep{xiao_smoothquant_2024,shao_omniquant_2024}, kronecker-factorized matrices \citep{sun_flatquant_2024}, or combinations thereof \citep{hu_ostquant_2025}. Each transform aids in recovering performance on quantized models; however, the literature remains scattered and no consensus on how to achieve an optimal transform has been reached.
The contributions of this paper are as follows:
\begin{itemize}
\item We introduce a novel framework to interpret the specific origins of quantization error, disentangling a concentration and an alignment term (Section~\ref{sec:framework}).
\item We demonstrate the effectiveness of our framework on LLM linear layers, analyzing how existing transforms improve quantization in terms of concentration and alignment. We further demonstrate that popular, rotation-based approaches (incl. Hadamard) completely neglect the alignment component of quantization error (Section~\ref{sec:analysis}). 
\item  We derive Concentration-Alignment Transform (CAT) as a theoretically motivated training-free transform that optimizes both components, but is too costly for practice. We show that approximating the CAT-optimal transform with a block-diagonal matrix (comparable in cost to existing solutions) still provides improvements in alignment and concentration, resulting in state-of-the-art accuracy on common LLM benchmarks (Sections~\ref{sec:cat} and \ref{sec:experiments}).
\end{itemize}

%% file: sections/method.tex
\section{Concentration-Alignment Framework}
\label{sec:framework}

For a quantized linear layer $\quant{\mW}\quant{\vx}$ with quantized weights $\quant{\mW}\defined Q_W(\mW)$ and quantized activations $\quant{\vx}\defined Q_x(\vx)$, the Signal to Quantized Noise Ratio (SQNR) on the layer output is defined as:
\begin{align}
    \SQNR{\quant{\mW}\quant{\vx}}\defined \frac{\E{\left\|\mW\vx\right\|_2^2}}{\E{\left\|\mW\vx-\quant{\mW}\quant{\vx}\right\|_2^2}}.
\end{align}
SQNR is a standard metric to measure the effect of the quantization noise since it measures the ratio of the magnitude of the signal (numerator) compared to the the quantization noise (denominator).

Using standard error de-correlation assumptions \citep{Widrow96}, the SQNR of a quantized linear layer can be approximated as a function of the SQNR obtained by quantizing only activation and weights, respectively. 
\begin{lemma}
\label{lemma:sqnr_composition}
    The SQNR of a quantized linear layer can be approximated with the harmonic sum (parallel) of the SQNR measured by quantizing activations and weights separately:
\begin{align*}
\SQNR{\quant{\mW}\quant{\vx}}\approx\SQNR{\mW \quant{\vx}}\parallel\SQNR{\quant{\mW}\vx},
\end{align*}
\end{lemma}
in which $\parallel$ refers to the \textit{parallel} operator: $a \parallel b = \left(1/a+1/b\right)^{-1}$.

Based on the independence of the components of the quantization error \citep{Gersho1977}, we show the following.

\begin{lemma}
\label{lemma:sqnr_activations}
Whenever the clipping error is small, the SQNR for uniform integer-quantized activations can be approximated as the product of three main components:
\begin{align*}
    \SQNR{\mW \quant{\vx}}\!\approx\! 12(\underbrace{2^{b_x}-1}_{N(b_x)})^2\underbrace{\frac{\E{\|\vx\|_2^2}}{\E{\range{\vx}^2}}}_{C(\vx)}\underbrace{\frac{\E{\|\mW\vx\|^2_2\|}}{\|\mW\|_F^2\E{\|\vx\|^2_2\|}}}_{A(\vx,\mW)},
\end{align*}
\end{lemma}
in which $b_x$ refers to the number of bits used to quantize the activations, $\|\mW\|_F^2$ indicates the squared Frobenius norm of the weights, and $\range{\vx}$ expresses  the quantization range of the activations, which may be \textit{static} (constant in $\vx$) or \textit{dynamic} (dependent on $\vx$). This is usually defined as $\max_i \vx_i-\min_i\vx_i$ for \textit{asymmetric} quantization and $2\max_i |\vx_i|$ for \textit{symmetric} quantization.

\begin{figure}[!tb]
    \centering
    \includegraphics[width=\linewidth]{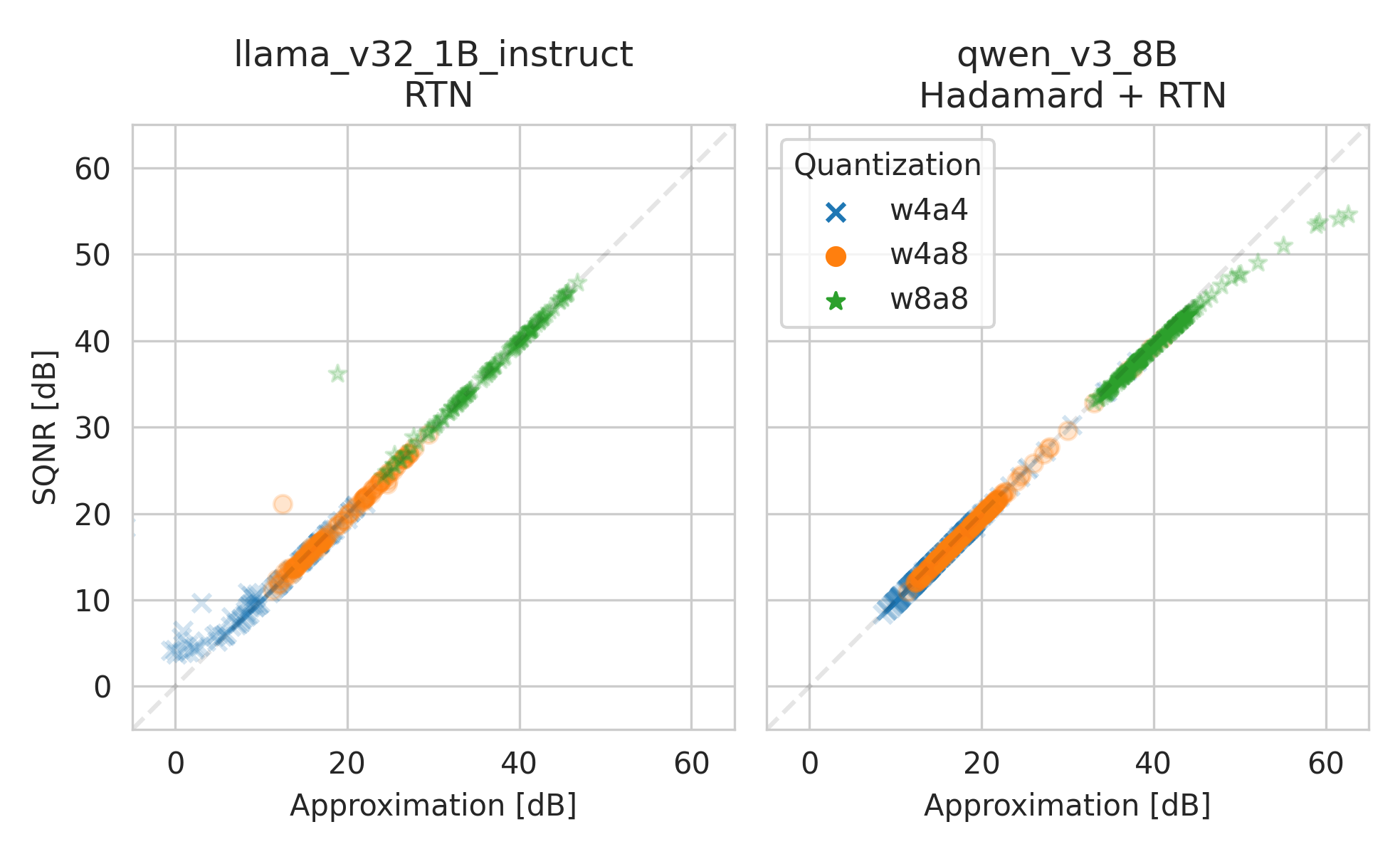}
    \caption{Empirical verification of the approximation reported in Theorem~\ref{th:main_result} for linear layers of LLama-v32-1B (left) and Qwen-v3-8B with Hadamard rotations applied before every linear layer (right) at W4A4, W4A8 and W8A8 quantization. Each dot represents one linear layer in the architecture. The approximation is close to the true SQNR for almost all layers. \textit{In L3.2 1B-it, the exception is \texttt{layer.1.mlp.down\_proj}, which is easier to quantize than the approximation suggests, due to the massive outlier of the [BOS] token \citep{sun_massive_2024} dominating the SQNR computation. For large SQNR, floating point issues limit the true SQNR (Qwen v3, top right)}.}
    \label{fig:approx_validity}
\end{figure}

A similar derivation can be used to approximate the SQNR for weight quantization.
\begin{lemma}
\label{lemma:sqnr_weights}
For negligible clipping error, the SQNR for uniform integer-quantized weights can be written as:
\begin{align*}
    \SQNR{\quant{\mW} \vx} \!\approx\! 12(\underbrace{2^{b_w}-1}_{N(b_w)})^2\!\underbrace{\frac{\sum_i\|\vw_i\|_2^2}{\sum_i\range{\vw_i}^2}}_{C(\mW)}\underbrace{\frac{\E{\|\mW\vx\|^2_2\|}}{\|\mW\|_F^2\E{\|\vx\|^2_2\|}}}_{A(\vx,\mW)},
\end{align*}
\end{lemma}
with $b_w$ as the number of bits used to quantize each element in a row of $\mW$.

Using the result from Lemma~\ref{lemma:sqnr_composition}, \ref{lemma:sqnr_activations}, and \ref{lemma:sqnr_weights}, we formulate an overall approximation for the SQNR of quantized linear layers.
\begin{theorem}
\label{th:main_result}
Whenever the clipping error is negligible, the SQNR for a quantized linear layer can be approximated as:
    \begin{align*}
    \!\!\!\SQNR{ \quant{\mW}\quant{\vx}}\!\approx\!12\!\left(N(b_x)^2C(\vx)\!\parallel\!N(b_w)^2C(\mW)\right)\!A(\vx, \mW).
\end{align*}
\end{theorem}

Figure~\ref{fig:approx_validity} empirically verifies the validity of the proposed approximation, comparing directly the measured SQNR and the value reported in Theorem~\ref{th:main_result} for linear layers of LLMs quantized at different bit widths with (right) and without (left) linear transformations. We observe that the reported approximation is accurate for the vast majority of layers between 5 and 50 decibels (dB), making it useful in practice.

Note that $N(b_x)$ and $N(b_w)$ indicate the number of quantization intervals, which depend solely on the chosen bit width. We will refer to $C(\vx)$ and $C(\mW)$ as measure of activation and weight \textit{Concentration}, respectively,
while we denote $A(\vx, \mW)$ as a measure of second-order activation and weight \textit{Alignment}.
In the following, we explore the role of each component in Theorem~\ref{th:main_result} in more depth.

\subsection{Interactions of Bit width, Concentration and Alignment}

\begin{figure}
    \centering
    \includegraphics[width=\linewidth, trim={0 0 0 0}, clip]{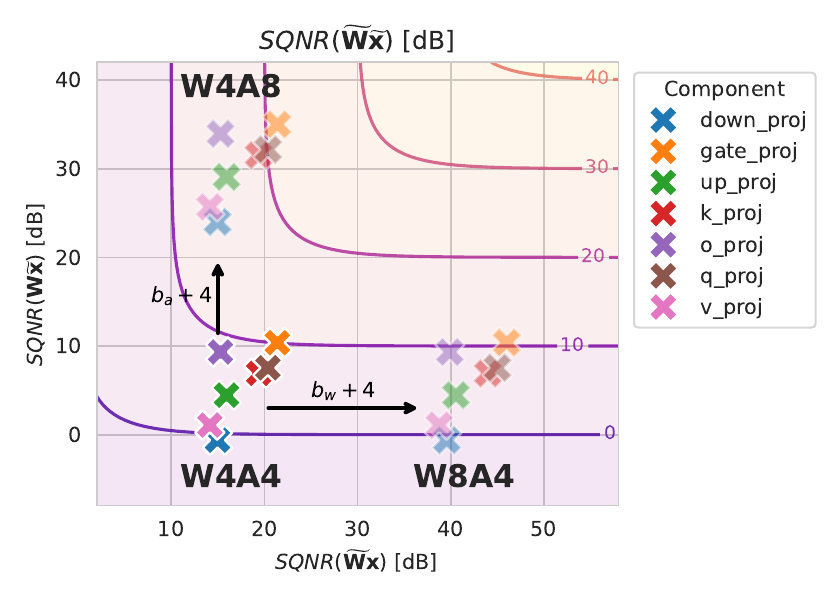}
    \caption{Comparison of activation quantization SQNR (y-axis), weight quantization SQNR (x-axis) and joint SQNR (iso-lines) for the linear layers of a Llama v3 8B architecture quantized at several bit widths. Starting from a 4 bit quntization of the linear layer (bottom left), increasing the weight bit width by 4 bits will result in a horizontal shift of around 24 dB, while increasing number of bits used for the activations results in a corresponding vertical shift. Since activation SQNR is worse than weight SQNR ($r(\vx,\mW)<1$), this latter scenario results in  much higher SQNR.}
    \label{fig:bit_shift}
\end{figure}

\paragraph{Bit width} The term $N(b_x)$ represents the number of distinct quantization intervals for $\vx$ (similarly, $N(b_w)$ for $\mW$). This term only depends on the number of bits used to quantize $\vx$ (and $\mW$). 
Higher bit width can be further used to compensate for poor weight/activation concentration. 
As illustrated in Figure~\ref{fig:bit_shift}, increasing the bit width of weights or activations, effectively increases overall SQNR. However, the magnitude of the increase depends on the ratio between $\SQNR{\mW\quant{\vx}}$ and $\SQNR{\mW\quant{\vx}}$, which corresponds to scenarios in which only activations or weights are quantized, respectively:
\begin{align}
    r(\vx,\mW) \defined \frac{\SQNR{\mW\quant{\vx}}}{\SQNR{\quant{\mW}\vx}} = \frac{b_x}{b_w}\frac{C(\vx)}{C(\mW)}.
\end{align}
Whenever the activation SQNR is smaller than the weight SQNR ($r(\vx,\mW)<1$), increasing the weight bit width $b_w$ will have less effect than increasing the activation bit width $b_x$, and vice versa for $r(\vx,\mW)>1$.
In other words, the overall SQNR for the linear layer is mostly determined by its worst component.

Whenever activations and weights are quantized at the same bit width $b_w=b_x=b$, the equation in Theorem~\ref{th:main_result} can be simplified to:
\begin{align}
    \SQNR{ \quant{\mW}\quant{\vx}}\approx\!12 N(b)^2\left(C(\vx)\!\parallel\!C(\mW)\right)\!A(\vx, \mW).
    \label{eq:main_result_same_bitwidth}
\end{align}
In this scenario, the effect of increasing the (joint) bit width $b$ can be disentangled from the other terms, and each extra bit increases joint SQNR approximately by a factor $4$ (about 6 dB). However, increasing bit width incurs extra memory and compute costs. For this reason, in this paper, we will mostly focus on improving concentration and alignment for a fixed bit width quantization.

\begin{figure*}[!t]
    \centering
    \begin{minipage}{0.465\textwidth}
        \centering
        \includegraphics[width=\textwidth, trim={0 0 4.5cm 0}, clip]{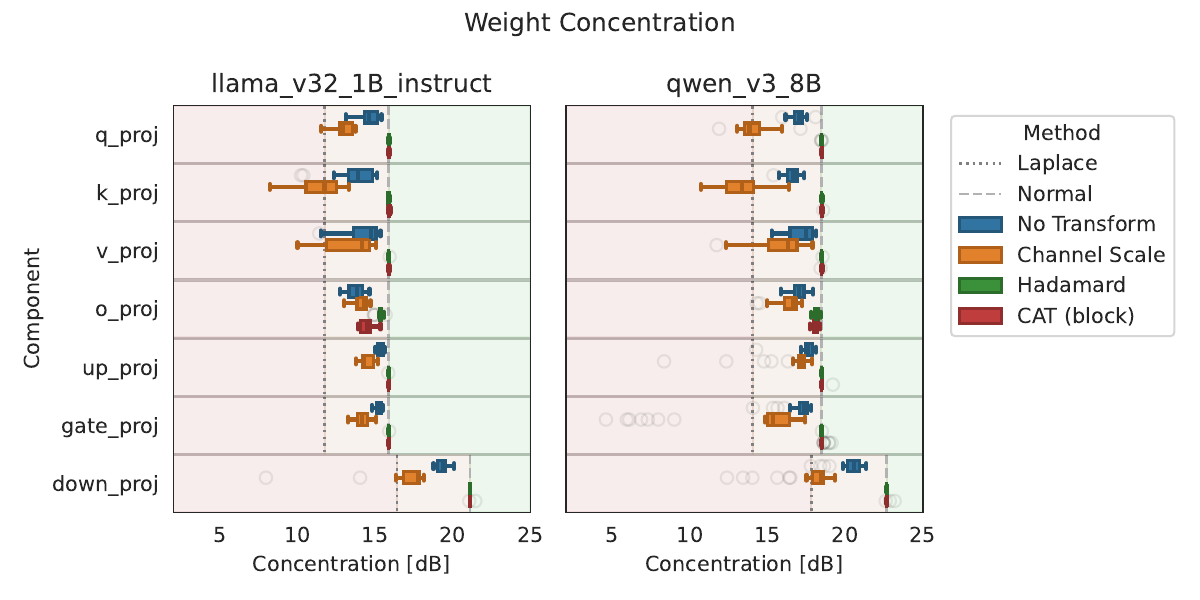}
    \end{minipage}
    \begin{minipage}{0.01\textwidth}
         ~
    \end{minipage}
    \begin{minipage}{0.51\textwidth}
        \centering\includegraphics[width=\textwidth, trim={2.7cm 0 0.3cm 0}, clip]{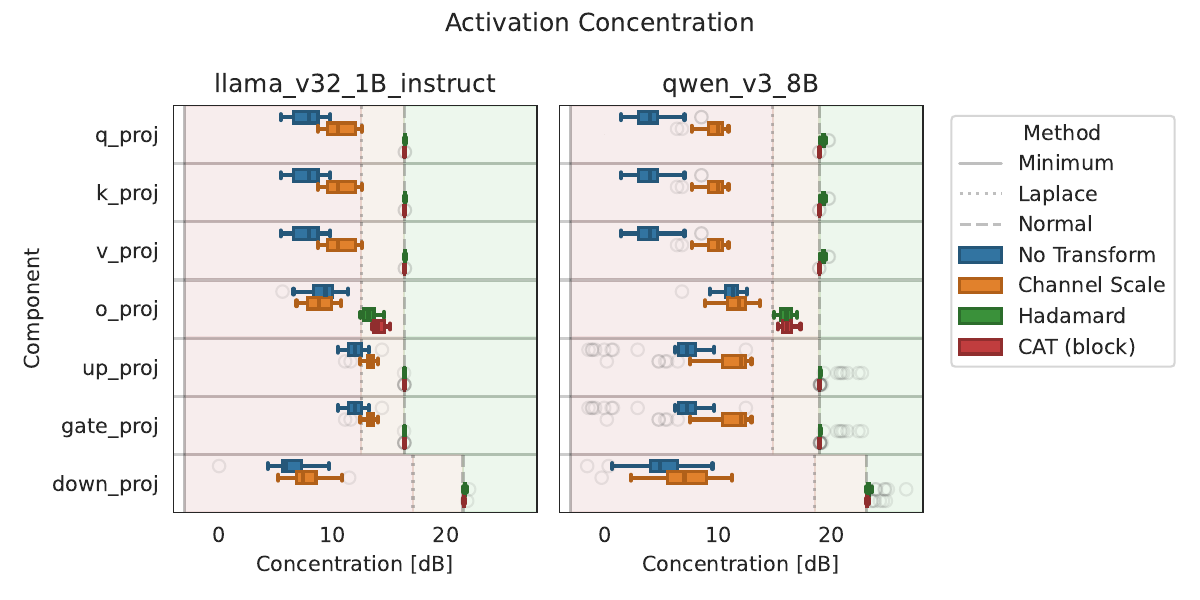}
    \end{minipage}
    \caption{Distribution of concentration of weight (left) and activation (right) quantization, for different layers and under different transforms. Activations are more heavy-tailed (worse than Laplace, red region) than weights, without transforms. Channel scaling (such as SmoothQuant) moves the activation outliers into the weights---this improves activation concentration, but worsens weight concentration significantly. Hadamard and CAT mix channels, which effectively makes them close to Gaussian for all layers. }
    \label{fig:analysis_concentration}
\end{figure*}

\paragraph{Concentration} $C(\mW)$  and $C(\vx)$ can be seen as measures of \textit{Concentration}, which defines the spread of the rows $\rw$  of the weights $\mW$ and activation distributions $\vx\sim p(\vx)$, respectively. Concentration is strongly related to the measure of kurtosis \citep{Sadegh25}, which considers the ratio between norm 4 and norm 2. Similarly, concentration captures the ratio between the squared norm 2 and squared norm infinity (range). Like kurtosis, concentration is scale-invariant and depends only on the weight and activation distributions.

For heavy-tailed distributions, in the presence of outliers, the concentration is low, while for distributions that collapse into a single value the concentration grows to infinity. For asymmetric quantization the lowest concentration is $1/2$ ($-3$ dB), while for symmetric quantization, the smallest value is $1/4$ ($-6$ dB), corresponding to a single non-zero value. Switching from a symmetric to an asymmetric quantization scheme can drastically reduce the quantization scale whenever the data is shifted or strongly asymmetric (e.g. activations after a $ReLU$ function), drastically improving concentration. However, asymmetric quantization yields only negligible improvements for symmetric, zero-centered distributions, which are common in modern LLMs.

Increasing activation or weight concentration has an analogous effect to increasing the corresponding bit width: whenever $r(\vx,\mW)<1$, as in the examples reported in Figure~\ref{fig:bit_shift} and Figure~\ref{fig:analysis_concentration}, improving activation concentration will have significantly more effect than improving the concentration of the corresponding weight.

\paragraph{Alignment} $A(\vx,\mW)$, on the other hand, can be interpreted as an \textit{Alignment} term, which measures the similarity between the directions of variation of the weights and activations.
Similar to concentration, alignment is also scale-invariant. Furthermore, alignment between activation and weights is not affected by rotations;
for any orthogonal matrix $\mR$:
\begin{align}
    A(\mR\vx,\mW\mR^{T}) =\frac{\E{\|\mW\mR^{T}\mR\vx\|^2_2\|}}{\|\mW\mR^{T}\|_F^2\E{\|\mR\vx\|^2_2\|}}= A(\vx,\mW).
\end{align}
In other words, rotating the  activations of a linear layer by $\mR$ and the corresponding weight by $\mR^T$ will have no effect on their alignment.

\begin{figure*}[!t]
    \centering
    \includegraphics[width=0.7\linewidth, trim={0 0 0cm 0}, clip]{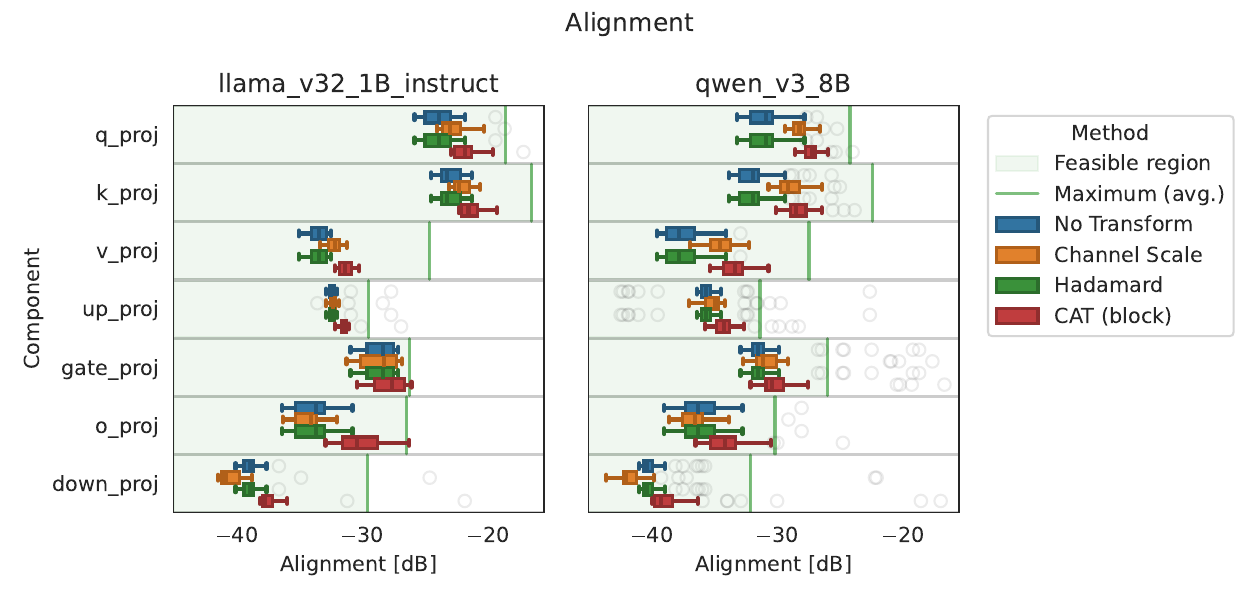}
    \caption{Distribution of alignment across layers under different transforms. The green shaded region indicates the achievable region, and underlines vast room for alignment improvements ( $ >10$ dB) on many layers. Note that rotation-based transforms cannot improve alignment, hence Hadamard transforms performs the same as no transforms. Channel scaling is similar to an alignment-optimal with block size 1. However, due to the sub-optimality scaling the channels using the ratio of the maximums does not necessarily improve on alignment on all layers. Block-diagonal matrices informed by CAT are good approximations of $\hat{\mM}$, and consistently improve alignment across all layers.}
    \label{fig:analysis_alignment}
\end{figure*}

It is important to realize that the alignment term $A(\vx, \mW)$ is a multiplier in each of the SQNR expressions. Therefore, improving alignment between activations and weights yields better SQNR for both weight and activation quantization, independent of the respective bit widths or concentration. Improving alignment by a factor $k$ has the same effect of increasing both activation and weight bit widths by approximately a factor $\log_2\sqrt{k}$. Across various LLMs we consistently observe that some layers (such as \textit{down\_proj}, \textit{o\_proj} and \textit{v\_proj}) display significantly worse alignment than the other layers (Figure~\ref{fig:analysis_alignment}). This raises the question: what is the best possible alignment? And how can we achieve it? In the next section, we answer this question by analyzing the impact of common linear transformations on weight and activation concentration and alignment.

\section{Analysis of Linear Transformations}
\label{sec:analysis}
Given a linear layer, we consider an invertible linear transformation represented by matrix $\mT$, which acts on activation and weights before quantizing:
\begin{align}
    \mW\vx = \mW\mT^{-1}\mT\vx \approx \underbrace{Q(\mW\mT^{-1})}_{\quant{\mW\mT^{-1}}}\underbrace{Q(\mT\vx)}_{\quant{\mT\vx}}.
\end{align}
Clearly, $\mT$ can be optimized to improve SQNR. In particular, it is easy to show that there are transformations for which the SQNR does not decrease:
\begin{align}
\SQNR{\quant{\mW}\quant{\vx}}\le \max_{\mT} \SQNR{\quant{\mW\mT^{-1}}\quant{\mT\vx}}.
\end{align}
Optimizing for $\mT$ directly can be challenging due to the discretization in the quantization function.
Instead, using the expression in Theorem~\ref{th:main_result}, we can interpret $\mT$ based on its effect on concentration and alignment.

\begin{figure*}[!t]
    \centering
    \includegraphics[width=0.9\linewidth, trim={0 0 0 0}, clip]{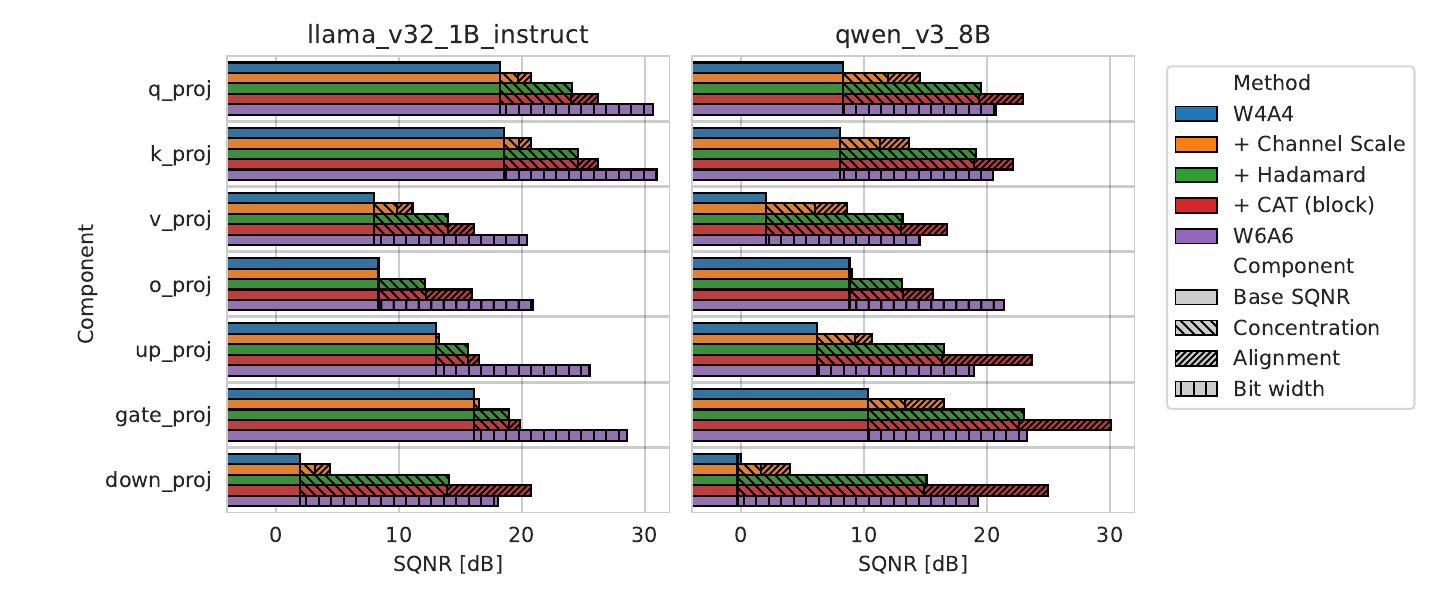}
    \caption{By designs transform for both alignment and concentration (CAT) drastically improve SQNR on all layers. In particular, Transformed W4A4 surpasses even W6A6 SQNR for all layers (except o\_proj) on the Qwen v3 8B architecture. Improvements are most notable for the MLP layers (gate, up, down), which have poor concentration and alignment without transforms.}
    \label{fig:analysis_sqnr}
\end{figure*}

In this section, we support our theoretical analysis with empirical observations collected by transforming linear layers of a \textit{Llama v3.2 1B} and \textit{Qwen v3} architectures. For each experiment, we apply transform $\mT$ before each linear layer and its inverse $\mT^{-1}$ on the weight of each linear layer. For layers that use the same input activations (such as \textit{q\_proj} / \textit{k\_proj} / \textit{v\_proj} and \textit{up\_proj} / \textit{gate\_proj}) we use the same transformation, treating the layer as a single linear layer with multiple output heads. The results are visualized in Figures ~\ref{fig:analysis_concentration}, \ref{fig:analysis_alignment} and \ref{fig:analysis_sqnr}.

\citet{xiao_smoothquant_2024} introduce channel-wise scaling $\mT_\text{channel}=\diag{[s_1,\hdots,s_d]}$, with $s_i = (\max_{\vx} x_i)^\alpha/(\max_{\vw_j} w_{ji})^{1-\alpha}$ to balance scale of activation and weight channels, effectively improves SQNR by re-distributing outliers between weights and activations. This operation has essentially two effects. First, the activation concentration is improved at the cost of lower weight concentration (Figure~\ref{fig:analysis_concentration}). As discussed in the previous section and visualized in Figure~\ref{fig:analysis_sqnr}, this effect is overall beneficial since the overall SQNR is mostly determined by the worse concentration. Secondly, this channel-balancing operation has a slightly positive effect of activation and weight alignment for most layers in the architecture (Figure~\ref{fig:analysis_alignment}) although the overall impact of the improved alignment on SQNR is not very significant. 

A common choice of transformation that reduces the effect of outliers is the family of Hadamard matrices $\mH$ \citep{ashkboos_quarot_2024, liu_spinquant_2024, hu_ostquant_2025,chee2024quip}.
Under the common assumption of channel independence, we can intuitively understand the effect of Hadamard rotations as the averaging of random variables, which causes the channel distribution to approach a Normal distribution by the central limit theorem. The concentration of Hadamard-transformed weights and activations approaches the concentration of a multivariate $d$-dimensional Normal distribution: $\mathcal{N}(\mathbf{0},\eye)$.
This is empirically visualized in Figure~\ref{fig:analysis_concentration}, where the Normal concentration is visualized with vertical dashed lines depending on the dimensionality of the layer.

The activation concentration for most activations in the model is worse than Laplace (red region), indicating the presence of heavy tails and severe outliers, while weight generally exhibit concentration in between the one of a Laplace and a Normal distribution (yellow region). In most layers, a Hadamard rotation suffices to significantly improve the concentration. This improvement is particularly evident for larger layers of bigger architectures, such as the \textit{down\_proj} layer in the Qwen v3 8B model, in which a Hadamard rotation results in an improvement in activation concentration of more than 10 decibels.
As show in Figure~\ref{fig:analysis_sqnr}, compared to channel scaling, Hadamard transformations have a much more pronounced effect on the overall SQNR, drastically improving both activation and weight concentration. 

However, any orthogonal linear transformation has no impact on activation and weight alignment.
In fact, while optimizing concentration (or similar proxies) is a common practice in literature, alignment is often neglected.
In particular, it is crucial to consider that alignment is \textbf{invariant to rotations}, hence approaches that rely solely on rotations $\mR$ yield no improvement on alignment, as demonstrated in Figure~\ref{fig:analysis_alignment}. Appendix~\ref{app:conc_all_per_layer} displays a more extensive visualization of the data reported in Figure~\ref{fig:analysis_concentration} and \ref{fig:analysis_alignment} by reporting the values for each architecture layer.

\section{Concentration-Alignment Transforms}
\label{sec:cat}
In this section we design transforms that effectively optimizes both concentration and alignment.
\paragraph{Optimizing alignment}
 In order to improve on alignment,
 we take a closer look to determine which transformations $\hat{\mM}$ are optimal for $A(\hat{\mM}\vx, \mW\hat{\mM})$. Interestingly, the maximization admits an analytical solution, which is determined by the matrix geometric mean \citep{pusz1975} of the inverse autocorrelation $\covariance\defined\E{\vx\vx^T}$ and the weight autocorrelation $\wcovariance\defined\mW^T\mW$:
\begin{align}
    \hat{\mM} &\defined \argmax_\mM A({\mM}\vx, \mW{\mM}^{-1}) =  \left(\wcovariance\# \covariance^{-1}\right)^{1/2} \nonumber\\
    &\text{with } \mA\# \mB \defined \mA^{1/2}\left(\mA^{-1/2}\mB\mA^{-1/2}\right)^{1/2}\mA^{1/2}.
    \label{eq:optimal_m}
\end{align}
Intuitively, $\hat{\mM}$ maps the direction of variations of activations and weights into the same space:
\begin{align}
    \hat{\mM}\covariance\hat{\mM} = \hat{\mM}^{-1}\wcovariance\hat{\mM}^{-1} = \left(\covariance^{-1/2}\wcovariance\covariance^{-1/2}\right)^{1/2}.
\end{align}
The maximum value of alignment is determined by distribution of the eigenvalues of the layer output:
\begin{align}
    A(\hat{\mM}\vx, \mW\hat{\mM}^{-1}) &= \frac{\trace{\covariance^{-1/2}\wcovariance\covariance^{-1/2}}}{\trace{\left(\covariance^{-1/2}\wcovariance\covariance^{-1/2}\right)^{1/2}}^2}\nonumber\\
    &= \frac{\sum_{i=1}^d \lambda_i^2}{\left(\sum_{i=1}^d \lambda_i\right)^2},
\end{align}
in which $\lambda_i$ refer to the eigenvalues of $\ycovariance\defined\mW\covariance\mW^T$.

As demonstrated in Figure~\ref{fig:analysis_alignment}, alignment between model weights and activation is usually quite poor when compared to the best achievable value, indicated with a vertical green line. In particular, in large layers such as \textit{down\_proj}, the alignment can be improved by about 10 dB, which would be roughly equivalent to the effect of increasing weight and activation by 2 bits each. 

\paragraph{Approximately optimal transform}
\label{sec:illustrative_example}
Using the insights gained from the observations on alignment and concentration, we can design an approximately optimal transform $\hat{T}$ following a 2-step procedure:
\begin{enumerate}
    \item We compute the invertible transformation $\hat{\mM}$ that maximizes alignment (equation~\ref{eq:optimal_m}).
    \item We compose $\hat{\mM}$ with a Hadamard matrix $\mH$ that maximizes the (joint) concentration. Note that this operation does not affect the alignment since alignment is rotation-invariant.
\end{enumerate}

Note that, in general, $\hat{\mM}$ is a full rank matrix, which is impractical for aligning weights and activations as it would require an expensive online matrix multiplication. In practice, $\hat{\mM}$ thus needs to be approximated by transforms that are cheaper, but which are not fully optimal for alignment. This yields numerous choices, e.g. diagonal \citep{xiao_smoothquant_2024}, Kronecker decompositions \citep{sun_flatquant_2024}, or block-diagonal matrices, possibly with permutations of channels \citep{lin_duquant_2024}. 

\paragraph{CAT Block Approximation} We provide a simple block-diagonal example as a case study, which we refer to as Concentration Alignment Transform \textit{CAT (block)} in this Section, and use this for the results in Section \ref{sec:experiments}.

Let us consider a simple block-diagonal approximation $\hat{M}_\text{block}^k = \diag{\left[\hat{M_1},\hdots,\hat{M_{d/k}}\right]}$ composed of $d/k$ blocks of size $k$, each optimized to maximize alignment on a subset of dimensions.
This results in an approximately-optimal transform:
\begin{align}
    \hat{\mT}_{\text{block}}^k \defined \mH\hat{\mM}_{\text{block}}^k .
    \label{eq:transform_def}
\end{align}

Figure~\ref{fig:analysis_concentration} and \ref{fig:analysis_alignment} report the values of concentration and alignment for $\hat{\mT}_{\text{block}}^k$ for $k=128$. In all cases, we observe that $\hat{\mT}_{\text{block}}^k$ yields better concentration and alignment values resulting in higher SQNR. Figure~\ref{fig:analysis_sqnr} demonstrates that this simple transformation yields significant alignment improvement when compared to simple Hadamard rotations while retaining similar concentration levels. This results in gains of up to 10 dB in layers such as \textit{gate\_proj} and \textit{down\_proj},  exceeding the SQNR measured on a models quantized at 6 bits precision (in purple) on the Qwen v3 architecure.

Note that for $k=1$, the expression for $\hat{\mM}_\text{block}^k$ simplifies to a diagonal matrix  $\hat{\mM}_\text{block}^1=\diag{\vm}$ in which each component is determined by the ratio between the expected squared weights and activations $m_i = \sqrt{\E{x_i^2}/\sum_j w_{ij}^2}$.

\paragraph{Computation and Memory Overhead} For dimension $d$, sequence length $n$, and $k$ blocks, the block matmul cost is $k$ times cheaper than a dense matrix multiplication: $O(nk(d/k)^2)=O(nd^2/k)$, versus $O(nd^2)$. The cross-block Hadamard transform costs only $O(nk\log k)$, which is negligible compared to the block matmul. The additional memory to store CAT is $O(d^2/k)$
, again $k$ times smaller than the corresponding dense matmul.
Practically, CAT (block) closely resembles the transform in FlatQuant, which uses two Kronecker products, each costing the same $O(nd^2/k)$ compute and slightly less memory $O(d^2/k^2)$. Due to the similarity in complexity, CAT (block) can thus be expected to achieve similar inference speed-ups. For a block size of 128 and 4096 features, the CAT block memory overhead is minor --- about 1\% of the linear-layer parameters.

%% file: sections/related_work.tex
\section{Related Work}

 Linear transformations are function-preserving operations~\citep{van2025fptquant} applied to the weights or activations of pretrained networks, such that the output of the network remains the same, yet the representation of the weights or distribution of the activations are more amenable to a downstream task. In particular, linear transformations are a popular for reducing outliers in LLMs and LVMs, and thereby improve quantization. 

\citet{xiao_smoothquant_2024} note that activation quantization is more challenging than weight quantization because activations contain more outliers. They introduce SmoothQuant, which shifts these problematic activation outliers into the weights by applying a per-channel scaling factor to the activations entering linear layers. 
\citet{chee2024quip} investigate transforms that mix channels, though only in the context of weight quantization. QuaRot~\citep{ashkboos_quarot_2024} popularized randomized Hadamard transforms (RHTs), demonstrating these effectively suppress outliers by spreading outlier channels across all other channels equally. SpinQuant~\citep{liu_spinquant_2024} shows that different RHTs (i.e. different seeds) yield widely varying performance, yet the random component of RHTs is discrete and cumbersome to optimize. To address this, they extend QuaRot by introducing two unconstrained rotation matrices trained to minimize the standard causal language-modeling loss. 

Importantly, these rotations are inserted so that they can be fused into the model weights after training, eliminating most runtime overhead. \citet{lin_duquant_2024} propose online rotations composed of fixed channel permutations and block-diagonal rotations. Note that by constraining FPTs to rotations, QuaRot, SpinQuant, and DuQuant can improve concentration, but not alignment. Newer works go beyond rotations. OSTQuant~\citep{hu_ostquant_2025} integrates both scaling vectors and rotations. 

Most recently, FlatQuant~\citep{sun_flatquant_2024} introduces FPTs based on matrix multiplications with a Kronecker product of two smaller, invertible matrices, and train these to minimize quantization error. Although these non-rotation methods are motivated by outlier reduction, their training loss (per-block/end-to-end quantization error) can inadvertently also improve alignment. Concurrent work \citep{chen2026wush} derives an expression for optimal linear transformations by focusing on low-precision Floating Point quantization.

%% file: sections/experiments.tex
\section{Experiments} \label{sec:experiments}

\begin{table*}[bt]
\renewcommand\arraystretch{1}
\centering
\caption{\small \textbf{A CAT block diagonal matrix outperforms baselines, and performs on par even without training.} Comparison of WikiText perplexity and averaged accuracy on six zero-shot common sense reasoning tasks. Average and standard deviation over 4 seeds.}
\label{tab:main}
\setlength{\tabcolsep}{1mm}
{\resizebox{\textwidth}{!}{
\input{tables/results}}}
\end{table*}

\paragraph{Models} We choose a range of models to evaluate all methods. We use Llama 2 7B~\citep{touvron_llama_2023}, Llama 3 8B \citep{grattafiori_llama_2024} to allow direct comparison to reported results from QuaRot, SpinQuant, and FlatQuant. We add to this Llama 3.2 1B instruct as a small model that is popular for edge devices. We also include Ministral 8B instruct~\citep{ministral} and Qwen 3 8B to cover other model families. We provide the code to reproduce the main experiments at \url{https://github.com/Qualcomm-AI-research/concentration_alignment_transforms}.

\paragraph{Calibration}
For all methods, we use 128 sequences of length 2048 of DCLM-edu for range setting and transform calibration/training. This is the same number as used in QuaRot and FlatQuant, though these related works calibrate on Wikitext. We prefer DCLM-edu, to ensure our Wikitext perplexity evaluation is not biased in favour of methods that may overfit the calibration data's characteristics (observed in some related works, e.g. \citep{van2025fptquant}).

\paragraph{Quantization set-up} We use the same set-up as used in QuaRot and follow-up works: linear layer inputs and weights in the transformer blocks and all KV cache is quantized. Activations and KV cache are quantized dynamically per token, asymmetrically, and weights are quantized per channel, symmetrically. For weight quantization, we run both round-to-nearest (RTN) and GPTQ (default settings) set-ups. We use $L_{2.4}$ for weight range estimation, following GPTQ.

\paragraph{Evaluation}
We evaluate Wikitext-2 perplexity at sequence length 2048~\citep{merity_pointer_2017}, and use LM-harness to evaluate the same common sense reasoning tasks used in FlatQuant~\citep{sun_flatquant_2024}: PIQA~\citep{bisk_piqa_2020}, WinoGrande~\citep{sakaguchi_winogrande_2021}, HellaSwag~\citep{zellers_hellaswag_2019}, ARC-e and ARC-c~\citep{clark_think_2018}, and LAMBADA~\citep{paperno_lambada_2016}.

\paragraph{Baselines} We compare block CAT (Section \ref{sec:illustrative_example}) with and without training, against the original floating point model (FP), PTQ using rounding-to-nearest (RTN), GPTQ \citep{frantar_gptq_2022}, prior rotation-based transforms QuaRot~\citep{ashkboos_quarot_2024} and SpinQuant~\citep{liu_spinquant_2024}, and the more expensive but state-of-the-art FlatQuant~\citep{sun_flatquant_2024}. To avoid differences in data, calibration, quantization, and evaluation set-up, we use the same pipeline for all methods.

\paragraph{Results} As reported in Table \ref{tab:main}, in the RTN setting, we observe that without training \textit{CAT (block)} outperforms all baselines on perplexity. On 0-shot performance, it performs comparable to the trained FlatQuant. With additional training, \textit{CAT (block)} improves further---especially on 0-shot performance, where it generally outperforms FlatQuant.

The GPTQ setting shows similar results, though now, baselines QuaRot and SmoothQuant have improved significantly compared to RTN, whereas GPTQ does not help \textit{CAT (block) w/ train} or FlatQuant substantially, consistently with the results reported in \citep{sun_flatquant_2024}. This is likely due to FlatQuant and CAT using learnable weight clipping during calibration, countering some of GPTQ's benefits.

Comparing our baselines to Llama 2 7B and Llama 3 8B results reported in \citep{sun_flatquant_2024}, our QuaRot is surprisingly better---likely due to switching to $L_{2.4}$ range setting, in line with other methods. SpinQuant GPTQ achieves similar results: our perplexity numbers are as expected slightly worse due to switching from Wikitext to DCLM-edu for calibration, but 0-shot results are comparable. Similarly, the reported results are comparable to the ones reported in FlatQuant although the method has been re-implemented from scratch.

Overall, we observe that the CAT calibration is not very sensitive to the size or choice of the calibration set although we observe minor improvements when using different splits from the same dataset for calibration and evaluation and higher variance. Results on several datasets and calibration sizes are reported in Appendix~\ref{app:calibration_ablation}.

\begin{figure*}[!t]
    \centering
    \includegraphics[width=\textwidth]{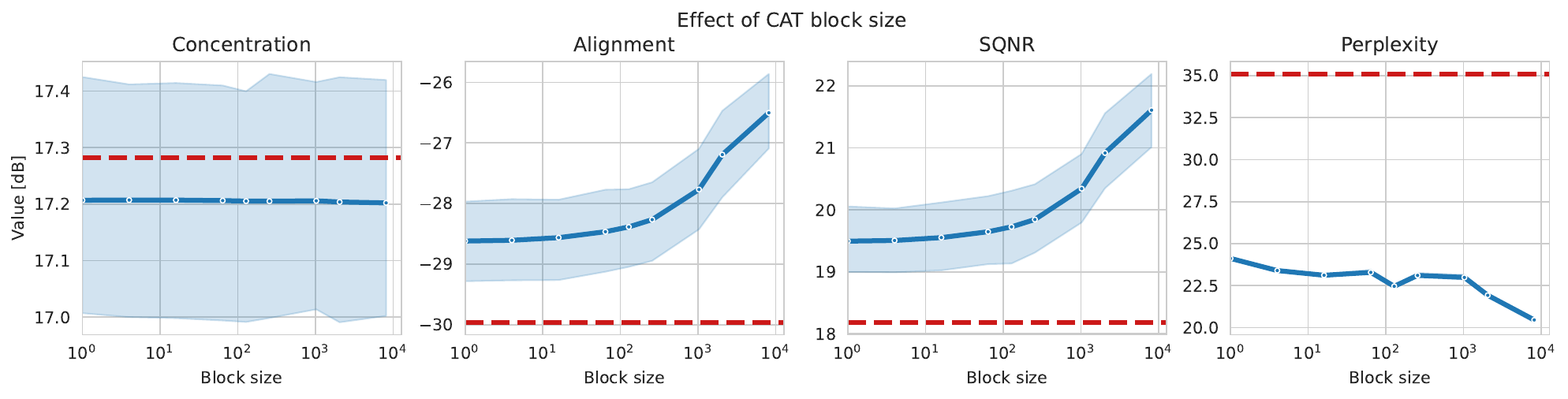}
    \caption{Effect of block size for $\hat{\mM}_{\text{block}}^k$ on the LLama 3.2 1B architecture ranging from 0 (Hadamard only, in dashed red lines) to full rank measured in average alignment, concentration, SQNR and Wikitext perplexity. The CAT block transform is not applied to the output projection of the attention layer since the corresponding block size is fixed and determined by the head dimension. The values of concentration, alignment and SQNR are average across 3 seeds and all model layers.}
    \label{fig:block_size}
    
\end{figure*}

\paragraph{Block size ablation} To further validate the effect of the alignment component $\hat{\mM}_{\text{block}}^k$, Figure~\ref{fig:block_size} the values of concentration, alignment, SQNR and perplexity for varying block size. Overall, we measure that larger block size results in higher alignment, while concentration remains unchanged due to the subsequent Hadamard matrix multiplication (Equation~\ref{eq:transform_def}). As a result, the local SQNR increases with block size and perplexity decreases, follows the same trend.

%% file: tables/results.tex
\begin{tabular}{llc|cc|cc|cc|cc|cc}
\toprule
 Weight & Transform & Train &\multicolumn{2}{c|}{\textbf{Llama 2 7B}} & \multicolumn{2}{c|}{\textbf{Llama 3 8B}} & \multicolumn{2}{c|}{\textbf{Llama 3.2 1B it}} & \multicolumn{2}{c|}{\textbf{Ministral 8B it}} & \multicolumn{2}{c}{\textbf{Qwen 3 8B}} \\
 quant-& method & & Wiki & 0-Shot$^6$ & Wiki & 0-Shot$^6$ & Wiki & 0-Shot$^6$ & Wiki & 0-Shot$^6$ & Wiki & 0-Shot \\
 ization   &     &       & ($\downarrow$)      & Avg.($\uparrow$)     & ($\downarrow$)        & Avg.($\uparrow$)      & ($\downarrow$)        & Avg.($\uparrow$)      & ($\downarrow$)        & Avg.($\uparrow$)     & ($\downarrow$)        & Avg.($\uparrow$)                 \\ \hline
 FP &  &  & 5.47 & 69.46 & 6.14  & 73.43  &  13.16 & 59.95  & 6.97 &  74.36 &  9.73 & 70.18 \\
\cline{1-13}
\multirow[m]{7}{*}{RTN} & None & \xmark & $1621.86^{\pm 0.00}$ & $31.51^{\pm 0.00}$ & $321.22^{\pm 0.00}$ & $35.31^{\pm 0.00}$ & $412.51^{\pm 0.00}$ & $32.18^{\pm 0.00}$ & $63.59^{\pm 0.00}$ & $38.81^{\pm 0.00}$ & $23254.67^{\pm 0.00}$ & $29.79^{\pm 0.00}$ \\
 & SmoothQuant & \xmark &$322.44^{\pm 38.39}$ & $34.49^{\pm 0.89}$ & $233.61^{\pm 6.24}$ & $35.19^{\pm 0.35}$ & $304.22^{\pm 11.86}$ & $32.87^{\pm 0.22}$ & $51.35^{\pm 1.38}$ & $41.58^{\pm 0.68}$ & $5047.04^{\pm 1171.81}$ & $30.22^{\pm 0.29}$ \\
 & QuaRot & \xmark & $9.31^{\pm 0.00}$ & $57.54^{\pm 0.00}$ & $10.99^{\pm 0.00}$ & $60.55^{\pm 0.00}$ & $29.89^{\pm 0.00}$ & $47.35^{\pm 0.00}$ & $9.12^{\pm 0.00}$ & $68.10^{\pm 0.00}$ & $13.69^{\pm 0.00}$ & $59.82^{\pm 0.00}$ \\
 & CAT (block) & \xmark & $\bf 6.11^{\pm 0.02}$ & $\bf 66.22^{\pm 0.58}$ & $\bf 8.29^{\pm 0.78}$	& $\bf 67.77^{\pm 0.33}$ & $\bf 18.25^{\pm 0.21}$ & $\bf 52.89^{\pm 0.21}$ & $\bf 8.07^{\pm 0.04}$ & $\bf 71.20^{\pm 0.35}$ & $\bf 10.82^{\pm 0.08}$ & $\bf 66.64^{\pm 0.32}$ \\
 \cline{2-13}
& SpinQuant & \cmark & $6.96^{\pm 0.03}$ & $63.64^{\pm 0.15}$ & $8.87^{\pm 0.08}$ & $66.33^{\pm 0.59}$ & $20.16^{\pm 0.06}$ & $51.45^{\pm 0.27}$ & $8.72^{\pm 0.04}$ & $70.30^{\pm 0.58}$ & $10.14^{\pm 0.03}$ & $65.26^{\pm 0.55}$ \\
 & FlatQuant & \cmark & $6.05^{\pm 0.01}$ & $66.67^{\pm 0.17}$ & $7.61^{\pm 0.01}$ & $68.34^{\pm 0.09}$ & $17.63^{\pm 0.22}$ & $54.14^{\pm 0.24}$ & $7.95^{\pm 0.03}$ & $71.86^{\pm 0.32}$ & $10.69^{\pm 0.04}$ & $66.25^{\pm 0.37}$ \\
 
 & CAT (block) & \cmark & $\bf 5.95^{\pm 0.01}$ & $\bf 67.34^{\pm 0.35}$ & $\bf 7.33^{\pm 0.01}$ & $\bf 69.66^{\pm 0.25}$ & $\bf 16.66^{\pm 0.20}$ & $\bf 54.74^{\pm 0.51}$ & $\bf 7.88^{\pm 0.08}$ & $\bf 71.99^{\pm 0.17}$ & $\bf 10.50^{\pm 0.02}$ & $\bf 66.93^{\pm 0.41}$ \\
\cline{1-13}
\multirow[m]{7}{*}{GPTQ} & None & \xmark & $1838.11^{\pm 145.28}$ & $31.88^{\pm 0.39}$ & $149.03^{\pm 10.56}$ & $36.77^{\pm 0.35}$ & $1514.48^{\pm 202.46}$ & $32.66^{\pm 0.54}$ & $30.62^{\pm 0.17}$ & $45.93^{\pm 0.62}$ & $19214.68^{\pm 857.18}$ & $29.93^{\pm 0.35}$ \\
  & SmoothQuant & \xmark & $40.06^{\pm 2.30}$ & $44.62^{\pm 1.66}$ & $109.26^{\pm 2.90}$ & $39.11^{\pm 1.53}$ & $760.78^{\pm 178.60}$ & $32.89^{\pm 0.63}$ & $24.02^{\pm 0.22}$ & $50.68^{\pm 0.45}$ & $3306.28^{\pm 340.65}$ & $30.00^{\pm 0.25}$ \\
 & QuaRot & \xmark & $6.22^{\pm 0.02}$ & $66.06^{\pm 0.27}$ & $8.21^{\pm 0.02}$ & $67.15^{\pm 0.19}$ & $19.97^{\pm 0.19}$ & $50.93^{\pm 0.30}$ & $8.18^{\pm 0.03}$ & $71.19^{\pm 0.47}$ & $11.50^{\pm 0.07}$ & $64.61^{\pm 0.53}$ \\
 & CAT (block) & \xmark & $\bf 6.01^{\pm 0.02}$ & $\bf 67.25^{\pm 0.14}$ & $\bf 7.60^{\pm 0.04}$ & $
 \bf 68.32^{\pm 0.49}$ & $\bf 17.89^{\pm 0.15}$ & $\bf 54.15^{\pm 0.53}$ & $\bf 7.94^{\pm 0.02}$ & $\bf 71.74^{\pm 0.11}$ & $\bf 10.58^{\pm 0.14}$ & $\bf 67.14^{\pm 0.68}$ \\
 \cline{2-13}
 & SpinQuant & \cmark & $6.26^{\pm 0.03}$ & $66.12^{\pm 0.38}$ & $7.88^{\pm 0.05}$ & $68.75^{\pm 0.41}$ & $18.16^{\pm 0.11}$ & $53.80^{\pm 0.16}$ & $8.28^{\pm 0.01}$ & $71.39^{\pm 0.15}$ & $9.94^{\pm 0.06}$ & $65.59^{\pm 0.42}$ \\
 & FlatQuant & \cmark & $\bf 5.94^{\pm 0.00}$ & $\bf67.58^{\pm 0.26}$ & $\bf 7.25^{\pm 0.02}$ & $\bf 69.97^{\pm 0.23}$ & $\bf 16.34^{\pm 0.23}$ & $\bf 55.35^{\pm 0.36}$ & $\bf7.72^{\pm 0.02}$ & $\bf72.31^{\pm 0.36}$ & $\bf 10.39^{\pm 0.04}$ & $\bf 67.88^{\pm 0.19}$ \\
 & CAT (block) & \cmark & $5.97^{\pm 0.02}$ & $67.17^{\pm 0.09}$ & $\bf 7.25^{\pm 0.01}$ & $\bf 70.38^{\pm 0.43}$ & $\bf 16.33^{\pm 0.07}$ & $\bf 55.39^{\pm 0.17}$ & $7.80^{\pm 0.05}$ & $\bf 72.26^{\pm 0.16}$ & $10.44^{\pm 0.04}$ & $\bf 67.83^{\pm 0.42}$ \\
\cline{1-13}
\bottomrule
\end{tabular}

%% file: sections/conclusions.tex
\section{Conclusion}
Low bit width quantization results in substantial errors, yet the community's understanding of this error is limited. Recently, linear transforms prior to quantization (e.g. Hadamard transform) have become popular approaches to reducing errors, motivated by the idea of mixing large outliers into other channels. By decomposing linear layer quantization error into \textit{Concentration} (a measure of outliers) and \textit{Alignment} (a measure of the alignment of the data and weight principal directions), we show that outliers form only half of the quantization error story. We propose CAT, which designs transforms that improve both concentration \textit{and} alignment.

\paragraph{Limitations} CAT provides the optimal transform for maximum alignment, but we note that this transform is a full-rank matrix that is costly and counterproductive for efficiency. This research does not prove what transforms may provide the best speed-accuracy trade-off in approximating the CAT optimal transform. Nonetheless, in our results we show that a simple block-diagonal transform (similar in cost to FlatQuant) that approximates CAT, yields W4A4 results that are competitive even without training. Future research will explore better approximations ---e.g. by adding mergeable rotations or permutations that can improve the block-diagonal approximation. Nonetheless, we believe that the proposed CAT framework provides actionable insights into effective design of linear transformations for effective low-precision quantization.

\paragraph{Additional Architectures} This work focuses on demonstrating the effectiveness of CAT on LLM architectures, however, our theoretical formulation is generally applicable to any linear layer. The study of concentration, alignment and effectiveness of block CAT transforms on other architectures is a clear direction for future work.

%% file: appendix/proofs.tex
\section{Proofs}
\label{app:proofs}

\subsection{SQNR of quantized weights and activations}

In order to demonstrate the statement reported in Lemma~\ref{lemma:sqnr_composition}, we expand the quantization error for quantized weights and activations:
\begin{align*}
    \E{\left\|\mW\vx-\quant{\mW}\quant{\vx}\right\|_2^2} &= \E{\left\|\mW\vx -(\mW-\Delta\mW)(\vx - \Delta\vx)\right\|_2^2}\\
    &=\E{\left\|\mW\Delta\vx+\Delta\mW\vx - \Delta\mW\Delta\vx\right\|_2^2}\\
    &=\E{\left\|\mW\Delta\vx\right\|_2^2+\left\|\Delta\mW\vx\right\|_2^2 +\left\| \Delta\mW\Delta\vx\right\|_2^2}\\
    &\ \ \ \ \ \ +2\E{ \trace{\mW\Delta\vx\vx^T\Delta\mW^T} - \trace{\mW\Delta\vx\Delta\vx^T\Delta\mW^T}-\trace{\Delta\mW\vx\Delta\vx^T\Delta\mW^T}}\\
    &=\E{\left\|\mW\Delta\vx\right\|_2^2}+\E{\left\|\Delta\mW\vx\right\|_2^2 }+\E{\left\| \Delta\mW\Delta\vx\right\|_2^2}\\
    &\ \ \ \ \ \ + 2\trace{\Delta\mW^T\mW\E{\Delta\vx\vx^T}} - 2\trace{\Delta\mW^T\mW\E{\Delta\vx\Delta\vx^T}}-2\trace{\Delta\mW^T\Delta\mW\E{\vx\Delta\vx^T}}\\
    &\approx \E{\left\|\mW\Delta\vx\right\|_2^2}+\E{\left\|\Delta\mW\vx\right\|_2^2 } = \E{\left\|\mW\vx - \mW\quant{\vx}\right\|_2^2}+\E{\left\|\mW\vx-\quant{\mW}\vx\right\|_2^2},
\end{align*}
in which $\Delta\vx\defined \vx-\quant{\vx}$ and $\Delta\mW\defined \mW-\quant{\mW}$.
This approximation is based on the following assumptions:
\begin{itemize}
    \item The activations $\vx$ and the quantization noise $\Delta\vx$ are uncorrelated \citep{Widrow96}:
    \begin{align*}
        \E{\Delta\vx\vx^T} \approx \mathbf{0}
    \end{align*}
    \item Similarly, the product between the weights $\mW$ and their quantization error $\Delta\mW$ is small:
    \begin{align*}
        \Delta\mW^T\mW \approx \mathbf{0}
    \end{align*}
    \item The magnitude of the product between the quantization noises is negligible:
    \begin{align*}
        \E{\left\| \Delta\mW\Delta\vx\right\|_2^2}\approx 0
    \end{align*}
\end{itemize}
Hence:
\begin{align}
\SQNR{\quant{\mW} \quant{\vx}} &= \frac{\E{\left\|\mW\vx\right\|_2^2}}{\E{\left\|\mW\vx-\quant{\mW}\quant{\vx}\right\|_2^2}}\nonumber\\
&\approx \frac{\E{\left\|\mW\vx\right\|_2^2}}{\E{\left\|\mW\vx-\mW\quant{\vx}\right\|_2^2 + \left\|\mW\vx-\quant{\mW}\vx\right\|_2^2}}\nonumber\\
&=\left(\frac{\E{\left\|\mW\vx-\mW\quant{\vx}\right\|_2^2} + \E{\left\|\mW\vx-\quant{\mW}\vx\right\|_2^2}}{\E{\left\|\mW\vx\right\|_2^2}}\right)^{-1}\nonumber\\
&= \left(\SQNR{\mW\quant{\vx}}^{-1}+\SQNR{\quant{\mW}\vx}^{-1} \right)^{-1}\nonumber\\
&= \SQNR{\mW\quant{\vx}}\parallel\SQNR{\quant{\mW}\vx}
\label{eq:sqnr_parallel_proof}
\end{align}

\subsection{From SQNR to Concentration and Alignment}

\subsubsection{Activation Quantization SQNR}
We can express the MSE for a linear weight with quantized activations as:
\begin{align*}
    \E{\left\|\mW\vx-\mW\quant{\vx}\right\|_2^2} &= \E{\left\|\mW\Delta\vx\right\|_2^2}\\
    &=\E{\trace{\mW\Delta\vx\Delta\vx^T\mW^T}}\\
    &= \trace{\mW^T\mW\E{\Delta\vx\Delta\vx^T}}.
\end{align*}
Consider the autocorrelation term $\E{\Delta\vx\Delta\vx^T}$.
We make the following assumptions
\begin{itemize}
    \item The quantization noise has zero expectation:
    \begin{align*}
        \E{\Delta\vx} \approx \mathbf{0}.
    \end{align*}
    \item The quantization noise for each channel is de-correlated:
    \begin{align*}
        \forall i\neq j\ \ \E{\Delta\vx_i\Delta\vx_j^T} \approx \E{\Delta\vx_i}\E{\Delta\vx_j^T}.
    \end{align*}
    \item The clipping error is negligible.
    For a given quantization bit width $b_x$,
    \begin{align*}
        \lim_{b_x\to\infty} \Delta\vx \approx \mathbf{0}
    \end{align*}
    \item All quantization intervals have the same size $s_x$, and the quantization noise is uniformly distributed in each interval:
    \begin{align}
        \forall i\ \ \ \Delta\vx_i\sim\uniform{-s_x/2,s_x/2},
    \end{align}
     with $s_x\defined \frac{\range{\vx}}{2^{b_x}-1}$ as the size of each quantization interval and $\range{\vx}$ as the full size of the quantized interval.
\end{itemize}
Given these assumptions, following classical work \citep{Gersho1977}, we have:
\begin{align*}
    \E{\Delta\vx_i\Delta\vx_j}\approx\begin{cases}
    \E{\Delta\vx_i}\E{\Delta\vx_j}\approx 0\ \ \ \ \forall i\neq j\\
    \E{\int_{-s/2}^{s/2} \frac{\Delta\vx_i^2 d\Delta\vx_i}{s}} = \E{\frac{s^2}{12}} \ \ \ \forall i=j
    \end{cases}.
\end{align*}
Hence:
\begin{align*}
    \E{\Delta\vx\Delta\vx^T}\approx \eye \frac{\E{\range{\vx}^2}}{12(2^{b_x}-1)^2}.
\end{align*}
Plugging this result in the previous expression:
\begin{align*}
    \E{\left\|\mW\vx-\mW\quant{\vx}\right\|_2^2} &\approx \frac{\E{\range{\vx}^2}}{12(2^{b_x}-1)^2}\trace{\mW^T\mW}\\
    &= \frac{1}{12(2^{b_x}-1)^2}\frac{\E{\range{\vx}^2}}{\E{\|\vx\|_2^2}}\|\mW\|_F^2\E{\|\vx\|_2^2}.
\end{align*}
Lastly, using this expression into the SQNR for quantized activations, we get:
\begin{align}
    \SQNR{\mW\quant{\vx}}\approx 12(2^{b_x}-1)^2 \underbrace{\frac{\E{\|\vx\|_2^2}}{\E{\range{\vx}^2}}}_{C(\vx)}\underbrace{\frac{\E{\|\mW\vx\|_2^2}}{\|\mW\|_F^2\E{\|\vx\|_2^2}}}_{A(\mW,\vx)}.
    \label{eq:proof_activation_sqnr}
\end{align}

\subsubsection{Weight Quantization SQNR}
The derivation for the SQNR for weight quantization is analogous to the one described for activation quantization. Here we underline the crucial steps
\begin{align*}
    \E{\left\|\mW\vx-\quant{\mW}\vx\right\|_2^2} &= \E{\sum_i^d\left\|\Delta\vw_i^T\vx\right\|_2^2}\\
    &=\trace{\sum_{i=1}^d\left(\Delta\vw_i\Delta\vw_i^T\right)\E{\vx\vx^T}}.
\end{align*}
Here, the assumptions on $\Delta\vw_i$ are analogous to the ones enumerated for $\Delta\vx$. Intuitively, whenever $\mW$ has a large number of rows $d$, we can treat the summation $\sum_{i=1}^d\left(\Delta\vw_i\Delta\vw_i^T\right)$ as a (discrete) expectation $d\ \E{\Delta\vw\Delta\vw^T}$ with $\Delta\vw\sim \discrete{\{\vw_i\}_{i=1}^{d}}{1/d}$. Therefore:
\begin{align*}
    \sum_i\Delta\vw_i\Delta\vw_i^T\approx \eye \frac{\sum_{i=1}^d\range{\vw_i}^2}{12(2^{b_w}-1)^2}.
\end{align*}
Plugging this result back into the previous expression:
\begin{align*}
    \E{\left\|\mW\vx-\quant{\mW}\vx\right\|_2^2} &= \frac{\sum_i\range{\vw_i}^2}{12(2^{b_w}-1)^2}\trace{\E{\vx\vx^T}}\\
    &= \frac{1}{12(2^{b_w}-1)^2}\frac{\sum_{i=1}^d\range{\vw_i}^2}{\sum_{i=1}^d\|\vw_i\|_2^2}\E{\|\vx\|_2^2}\|\mW\|_F^2,
\end{align*}
hence:
\begin{align}
    \SQNR{\quant{\mW}\vx}\approx 12(2^{b_w}-1)^2 \underbrace{\frac{\sum_{i=1}^d\range{\vw_i}^2}{\sum_{i=1}^d\|\vw_i\|_2^2}}_{C(\mW)}\underbrace{\frac{\E{\|\mW\vx\|_2^2}}{\|\mW\|_F^2\E{\|\vx\|_2^2}}}_{A(\mW,\vx)}.
\label{eq:proof_weight_sqnr}
\end{align}

\subsection{Full SQNR expression}
We can combine the results from Equations~\ref{eq:sqnr_parallel_proof}, \ref{eq:proof_activation_sqnr}, and \ref{eq:proof_weight_sqnr} to get a final expression:
\begin{align}
    \SQNR{\quant{\mW} \quant{\vx}} &\approx \SQNR{\mW\quant{\vx}}\parallel\SQNR{\quant{\mW}\vx}\nonumber\\
    &\approx 12\left((2^{b_x}-1)^2 C(\vx)\parallel(2^{b_w}-1)^2 C(\mW)\right) A(\vx,\mW)\nonumber\\
\end{align}

%% file: appendix/additional_results.tex
\section{Additional Results}
\label{app:additional_results}

\subsection{Per-layer concentration and alignment}
\label{app:conc_all_per_layer}

\begin{figure}
    \centering
    \includegraphics[width=\linewidth]{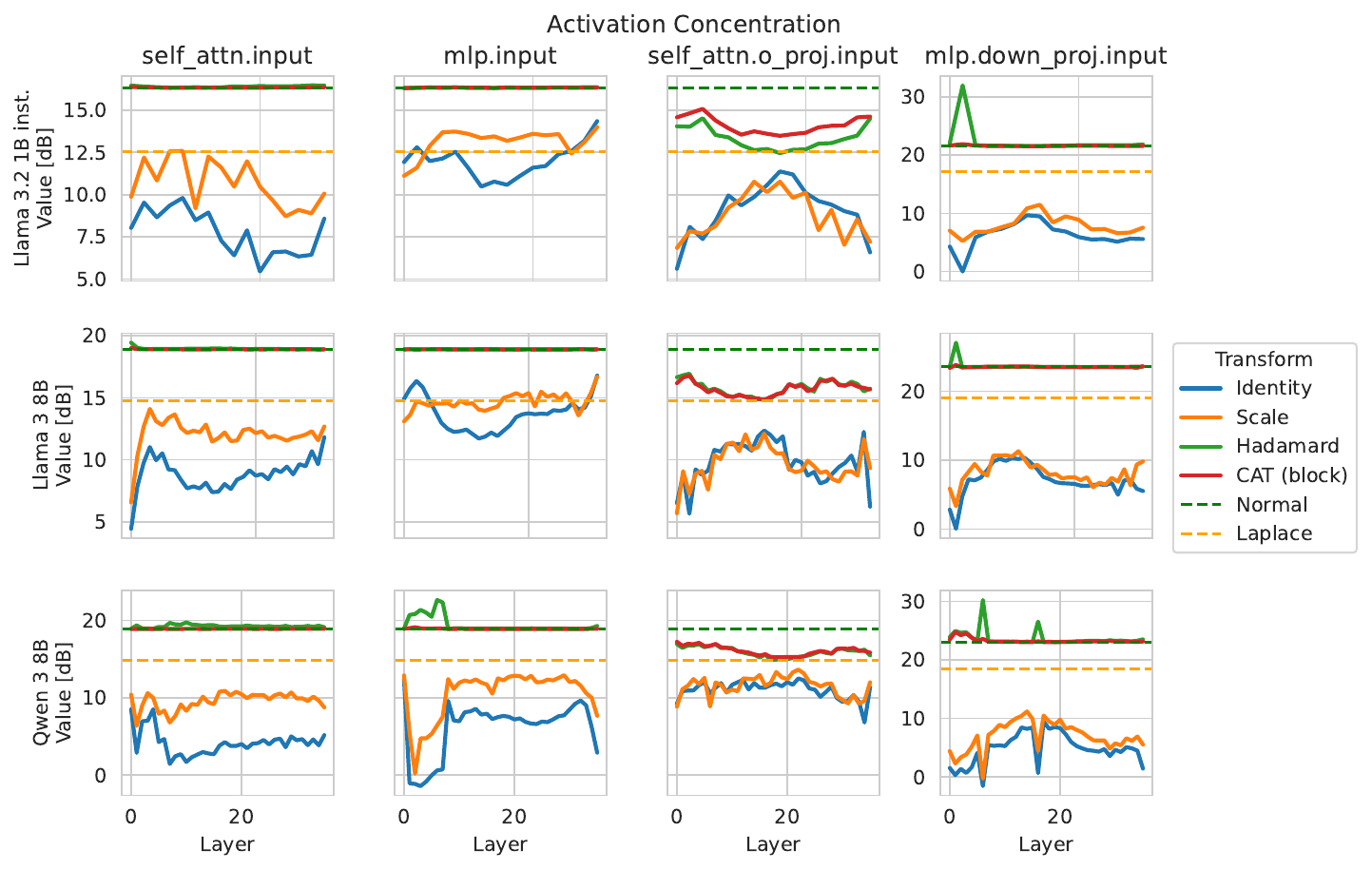}
    \caption{Concentration measured for input activations of all layers of three LLM architectures for 4 different transforms. Dashed lines indicate the reference concentration for a Normal (green) and Laplace distribution (orange).}
    \label{fig:per_layer_act_concentration}
\end{figure}

\begin{figure}
    \centering
    \includegraphics[width=\linewidth]{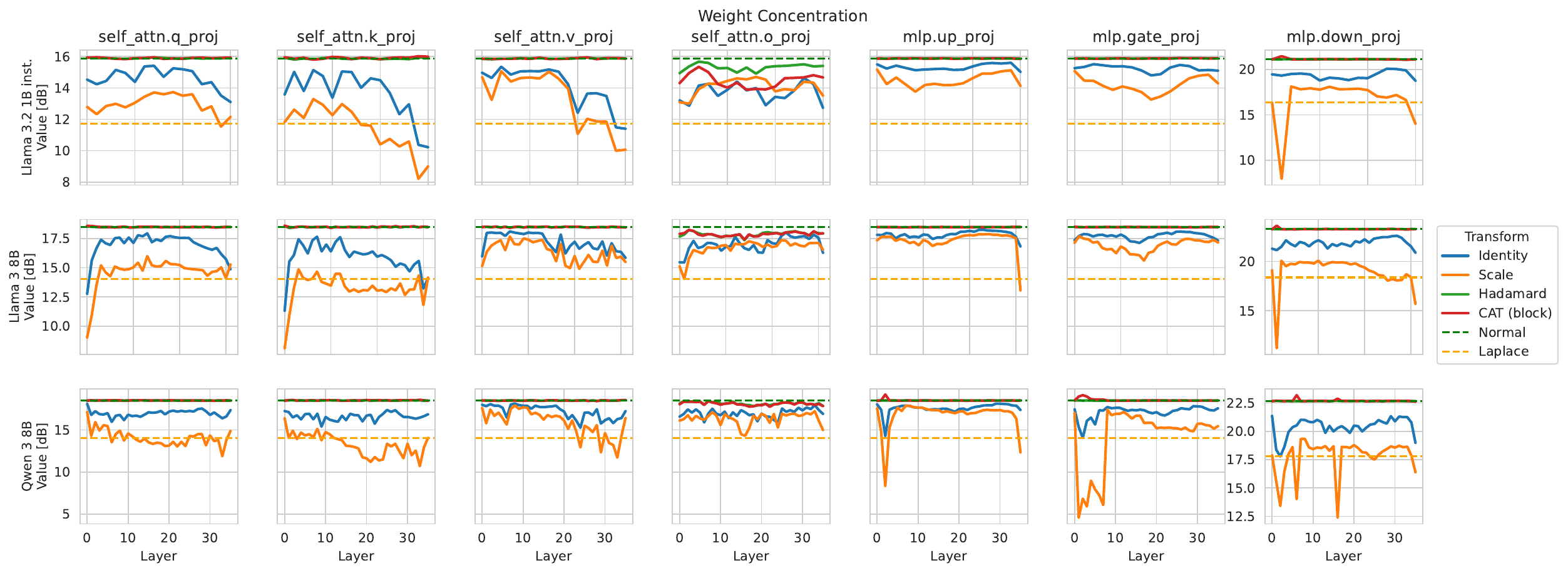}
    \caption{Concentration measured for weights of all layers of three LLM architectures for 4 different transforms. Dashed lines indicate the reference concentration for a Normal (green) and Laplace distribution (orange).}
    \label{fig:per_layer_weight_concentration}
\end{figure}

\begin{figure}
    \centering
    \includegraphics[width=\linewidth]{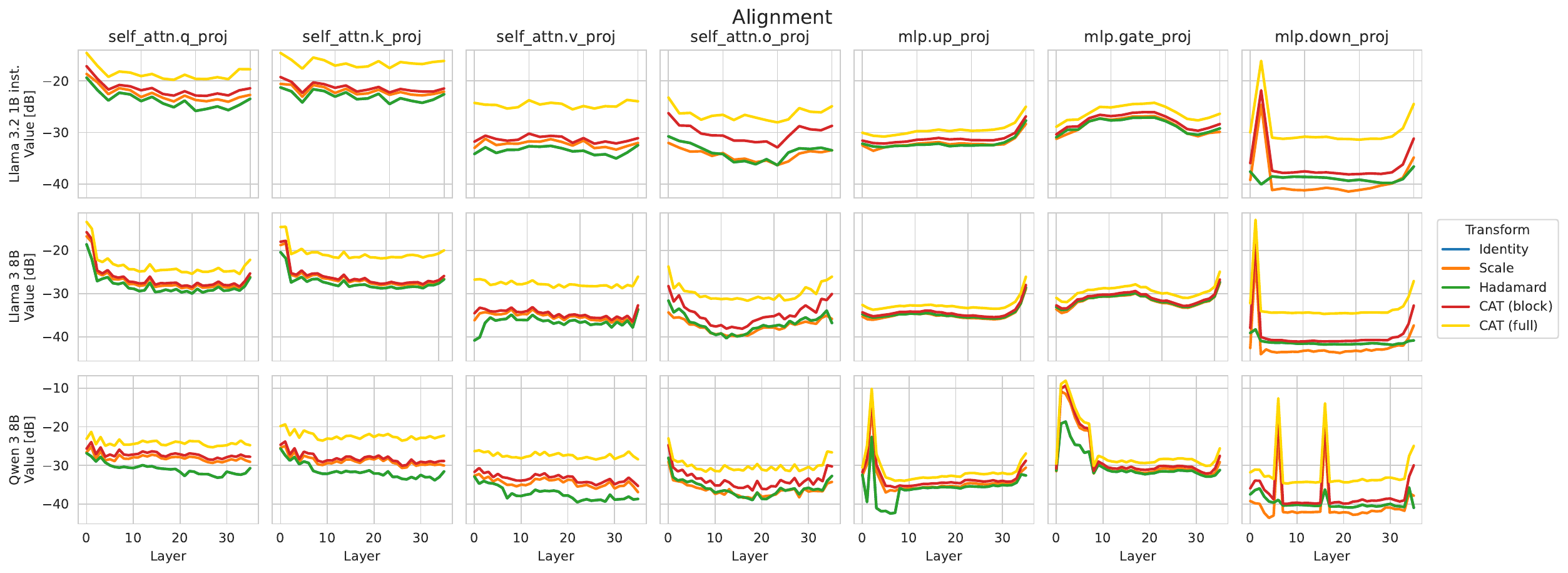}
    \caption{Alignment measured for all linear layers of three LLM architectures for 4 different transforms. The golden line represent the (optimal) alignment achieved with full rank CAT.}
    \label{fig:per_layer_alignment}
\end{figure}

Figure~\ref{fig:per_layer_act_concentration} and Figure~\ref{fig:per_layer_weight_concentration} report the values of concentration for each layer and layer type across the LLama 3.2 1B instruct, LLama 3 8B and, Qwen 3 8B architectures. The experimental setting correspond to the same setting used to produce Figure~\ref{fig:analysis_concentration}. Similarly, Figure~\ref{fig:per_layer_alignment} refers to the values of alignment for each layer of the three architectures as in Figure~\ref{fig:analysis_alignment}.

\subsection{Calibration Ablation}
\label{app:calibration_ablation}
\begin{figure}[!tbh]
    \centering
    \begin{minipage}{0.6\linewidth}
        \centering
        \caption{Effect of calibration set size (DCLM-edu) on CAT block performance for a Llama 3.2 1B instruct model. Values of concentration, alignment, and SQNR are averaged across the entire network and over three seeds. Quantization uses 4 bits for activations and weights. Quantization ranges are estimated using \textit{minmax}.}
        \label{tab:calibration_size}
        \vspace{1cm}
        {
        \setlength{\tabcolsep}{4pt} 
        \begin{tabular}{l|cccc}
            \toprule
            Size & 32 & 128 & 1024 & 4096 \\
            \midrule
            Conc. & $16.7 \pm 0.0$ & $16.7 \pm 0.0$ & $16.7 \pm 0.0$ & $16.7 \pm 0.0$ \\
            Align. & $-29.2 \pm 0.0$ & $-29.1 \pm 0.0$ & $-29.1 \pm 0.0$ & $-29.1 \pm 0.0$ \\
            SQNR & $18.6 \pm 0.0$ & $18.6 \pm 0.0$ & $18.6 \pm 0.0$ & $18.6 \pm 0.0$ \\
            PPL & $22.0 \pm 0.4$ & $21.7 \pm 0.3$ & $21.7 \pm 0.2$ & $22.1 \pm 0.3$ \\
            \cline{1-5}
            \bottomrule
        \end{tabular}
        }
    \end{minipage}
    \begin{minipage}{0.3\linewidth}
        \centering
        \includegraphics[width=\linewidth]{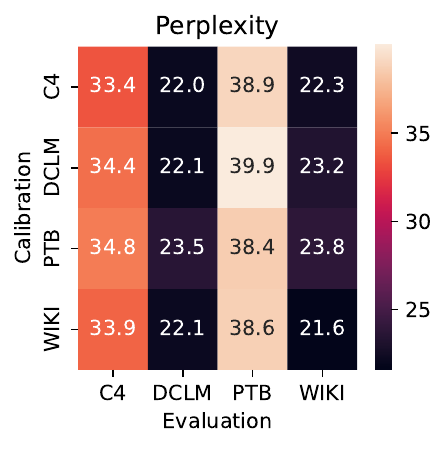}
        \caption{Performance of CAT blocks on W4A4 Llama 3.2 1B instruct for different choices of calibration and evaluation datasets (C4, WikiText, Penn Treebank, and DCLM-edu). The choice of calibration data has a limited impact on model perplexity.}
        \label{fig:calibration_data}
    \end{minipage}
\end{figure}

In order to assess the sensitivity of CAT to the choice of calibration set, we perform two ablation experiments.
Table~\ref{tab:calibration_size} evaluates the sensitivity to the calibration set size. Overall, we observe that the variation obtained when considering between 32 and 4096 sequences lies within the standard deviation observed when estimating CAT statistics across different seeds. Similarly, Figure~\ref{fig:calibration_data} shows the effect of selecting different datasets (each consisting of up to 4096 sequences) for calibration and evaluation, including \href{https://huggingface.co/datasets/allenai/c4}{C4}, \href{https://huggingface.co/datasets/Salesforce/wikitext}{WikiText}, \href{https://huggingface.co/datasets/ptb-text-only/ptb_text_only}{Penn Treebank}, and \href{https://huggingface.co/datasets/HuggingFaceTB/dclm-edu}{DCLM-edu}. The limited impact on the final perplexity supports the robustness of our method.

%% file: bibliography.bib
@inproceedings{hu_ostquant_2025,
	title = {{OSTQuant}: {Refining} {Large} {Language} {Model} {Quantization} with {Orthogonal} and {Scaling} {Transformations} for {Better} {Distribution} {Fitting}},
	url = {https://openreview.net/forum?id=rAcgDBdKnP},
	booktitle = {The {Thirteenth} {International} {Conference} on {Learning} {Representations}},
	author = {Hu, Xing and Cheng, Yuan and Yang, Dawei and Chen, Zhixuan and Xu, Zukang and {JiangyongYu} and {XUCHEN} and Yuan, Zhihang and jiang, Zhe and Zhou, Sifan},
	year = {2025},
}

@article{touvron_llama_2023,
  title={Llama 2: Open foundation and fine-tuned chat models},
  author={Touvron, Hugo and Martin, Louis and Stone, Kevin and Albert, Peter and Almahairi, Amjad and Babaei, Yasmine and Bashlykov, Nikolay and Batra, Soumya and Bhargava, Prajjwal and Bhosale, Shruti and others},
  journal={arXiv preprint arXiv:2307.09288},
  year={2023}
}

@article{grattafiori_llama_2024,
  title={The llama 3 herd of models},
  author={Grattafiori, Aaron and Dubey, Abhimanyu and Jauhri, Abhinav and Pandey, Abhinav and Kadian, Abhishek and Al-Dahle, Ahmad and Letman, Aiesha and Mathur, Akhil and Schelten, Alan and Vaughan, Alex and others},
  journal={arXiv preprint arXiv:2407.21783},
  year={2024}
}

@inproceedings{paperno_lambada_2016,
  title={The LAMBADA dataset: Word prediction requiring a broad discourse context},
  author={Paperno, Denis and Kruszewski, Germ{\'a}n and Lazaridou, Angeliki and Pham, Ngoc-Quan and Bernardi, Raffaella and Pezzelle, Sandro and Baroni, Marco and Boleda, Gemma and Fern{\'a}ndez, Raquel},
  booktitle={Proceedings of the 54th Annual Meeting of the Association for Computational Linguistics (Volume 1: Long Papers)},
  pages={1525--1534},
  year={2016}
}

@inproceedings{
merity_pointer_2017,
title={Pointer Sentinel Mixture Models},
author={Stephen Merity and Caiming Xiong and James Bradbury and Richard Socher},
booktitle={International Conference on Learning Representations},
year={2017},
}

@article{frantar_gptq_2022,
  title={Gptq: Accurate post-training quantization for generative pre-trained transformers},
  author={Frantar, Elias and Ashkboos, Saleh and Hoefler, Torsten and Alistarh, Dan},
  journal={arXiv preprint arXiv:2210.17323},
  year={2022}
}

@article{bisk_piqa_2020,
	title = {{PIQA}: {Reasoning} about {Physical} {Commonsense} in {Natural} {Language}},
	volume = {34},
	copyright = {Copyright (c) 2020 Association for the Advancement of Artificial Intelligence},
	issn = {2374-3468},
	shorttitle = {{PIQA}},
	url = {https://ojs.aaai.org/index.php/AAAI/article/view/6239},
	doi = {10.1609/aaai.v34i05.6239},
	language = {en},
	number = {05},
	urldate = {2024-12-17},
	journal = {Proceedings of the AAAI Conference on Artificial Intelligence},
	author = {Bisk, Yonatan and Zellers, Rowan and Bras, Ronan Le and Gao, Jianfeng and Choi, Yejin},
	month = apr,
	year = {2020},
	note = {Number: 05},
	pages = {7432--7439},
}

@article{sakaguchi_winogrande_2021,
	title = {{WinoGrande}: an adversarial winograd schema challenge at scale},
	volume = {64},
	issn = {0001-0782},
	shorttitle = {{WinoGrande}},
	url = {https://dl.acm.org/doi/10.1145/3474381},
	doi = {10.1145/3474381},
	number = {9},
	urldate = {2024-12-17},
	journal = {Commun. ACM},
	author = {Sakaguchi, Keisuke and Bras, Ronan Le and Bhagavatula, Chandra and Choi, Yejin},
	month = aug,
	year = {2021},
	pages = {99--106},
}

@misc{clark_think_2018,
	title = {Think you have {Solved} {Question} {Answering}? {Try} {ARC}, the {AI2} {Reasoning} {Challenge}},
	shorttitle = {Think you have {Solved} {Question} {Answering}?},
	url = {http://arxiv.org/abs/1803.05457},
	doi = {10.48550/arXiv.1803.05457},
	urldate = {2024-12-17},
	publisher = {arXiv},
	author = {Clark, Peter and Cowhey, Isaac and Etzioni, Oren and Khot, Tushar and Sabharwal, Ashish and Schoenick, Carissa and Tafjord, Oyvind},
	month = mar,
	year = {2018},
	note = {arXiv:1803.05457 [cs]},
}

@misc{zellers_hellaswag_2019,
	title = {{HellaSwag}: {Can} a {Machine} {Really} {Finish} {Your} {Sentence}?},
	shorttitle = {{HellaSwag}},
	url = {http://arxiv.org/abs/1905.07830},
	doi = {10.48550/arXiv.1905.07830},
	urldate = {2024-12-17},
	publisher = {arXiv},
	author = {Zellers, Rowan and Holtzman, Ari and Bisk, Yonatan and Farhadi, Ali and Choi, Yejin},
	month = may,
	year = {2019},
	note = {arXiv:1905.07830 [cs]},
}

@inproceedings{lin_duquant_2024,
	title = {{DuQuant}: {Distributing} {Outliers} via {Dual} {Transformation} {Makes} {Stronger} {Quantized} {LLMs}},
	shorttitle = {{DuQuant}},
	url = {http://arxiv.org/abs/2406.01721},
	doi = {10.48550/arXiv.2406.01721},
	urldate = {2024-11-15},
	booktitle = {Advances in {Neural} {Information} {Processing} {Systems}},
	publisher = {arXiv},
	author = {Lin, Haokun and Xu, Haobo and Wu, Yichen and Cui, Jingzhi and Zhang, Yingtao and Mou, Linzhan and Song, Linqi and Sun, Zhenan and Wei, Ying},
	month = nov,
	year = {2024},
	note = {arXiv:2406.01721},
}

@inproceedings{
sun_flatquant_2024,
title={FlatQuant: Flatness Matters for {LLM} Quantization},
author={Yuxuan Sun and Ruikang Liu and Haoli Bai and Han Bao and Kang Zhao and Yuening Li and JiaxinHu and Xianzhi Yu and Lu Hou and Chun Yuan and Xin Jiang and Wulong Liu and Jun Yao},
booktitle={Forty-second International Conference on Machine Learning},
year={2025},
url={https://openreview.net/forum?id=uTz2Utym5n}
}

@misc{ashkboos_quarot_2024,
	title = {{QuaRot}: {Outlier}-{Free} 4-{Bit} {Inference} in {Rotated} {LLMs}},
	shorttitle = {{QuaRot}},
	url = {https://arxiv.org/abs/2404.00456v1},
	language = {en},
	urldate = {2024-10-03},
	journal = {arXiv.org},
	author = {Ashkboos, Saleh and Mohtashami, Amirkeivan and Croci, Maximilian L. and Li, Bo and Jaggi, Martin and Alistarh, Dan and Hoefler, Torsten and Hensman, James},
	month = mar,
	year = {2024},
}

@misc{shao_omniquant_2024,
	title = {{OmniQuant}: {Omnidirectionally} {Calibrated} {Quantization} for {Large} {Language} {Models}},
	shorttitle = {{OmniQuant}},
	url = {http://arxiv.org/abs/2308.13137},
	doi = {10.48550/arXiv.2308.13137},
	urldate = {2024-10-07},
	publisher = {arXiv},
	author = {Shao, Wenqi and Chen, Mengzhao and Zhang, Zhaoyang and Xu, Peng and Zhao, Lirui and Li, Zhiqian and Zhang, Kaipeng and Gao, Peng and Qiao, Yu and Luo, Ping},
	month = mar,
	year = {2024},
	note = {arXiv:2308.13137 [cs]},
}

@misc{liu_spinquant_2024,
	title = {{SpinQuant}: {LLM} quantization with learned rotations},
	shorttitle = {{SpinQuant}},
	url = {https://arxiv.org/abs/2405.16406v2},
	language = {en},
	urldate = {2024-10-04},
	journal = {arXiv.org},
	author = {Liu, Zechun and Zhao, Changsheng and Fedorov, Igor and Soran, Bilge and Choudhary, Dhruv and Krishnamoorthi, Raghuraman and Chandra, Vikas and Tian, Yuandong and Blankevoort, Tijmen},
	month = may,
	year = {2024},
}

@article{sun_massive_2024,
  title={Massive Activations in Large Language Models},
  author={Sun, Mingjie and Chen, Xinlei and Kolter, J Zico and Liu, Zhuang},
  journal={arXiv preprint arXiv:2402.17762},
  year={2024}
}

@misc{tseng_quip_2024,
	title = {{QuIP}\#: {Even} {Better} {LLM} {Quantization} with {Hadamard} {Incoherence} and {Lattice} {Codebooks}},
	shorttitle = {{QuIP}\#},
	url = {https://arxiv.org/abs/2402.04396v2},
	language = {en},
	urldate = {2024-10-03},
	journal = {arXiv.org},
	author = {Tseng, Albert and Chee, Jerry and Sun, Qingyao and Kuleshov, Volodymyr and De Sa, Christopher},
	month = feb,
	year = {2024},
}

@misc{xiao_smoothquant_2024,
	title = {{SmoothQuant}: {Accurate} and {Efficient} {Post}-{Training} {Quantization} for {Large} {Language} {Models}},
	shorttitle = {{SmoothQuant}},
	url = {http://arxiv.org/abs/2211.10438},
	doi = {10.48550/arXiv.2211.10438},
	urldate = {2024-10-03},
	publisher = {arXiv},
	author = {Xiao, Guangxuan and Lin, Ji and Seznec, Mickael and Wu, Hao and Demouth, Julien and Han, Song},
	month = mar,
	year = {2024},
	note = {arXiv:2211.10438 [cs]},
}

@article{chee2024quip,
  title={Quip: 2-bit quantization of large language models with guarantees},
  author={Chee, Jerry and Cai, Yaohui and Kuleshov, Volodymyr and De Sa, Christopher M},
  journal={Advances in Neural Information Processing Systems},
  volume={36},
  year={2024}
}

@misc{ministral,
  title={Un Ministral, des Ministraux},
  author={Mistral.ai},
  year={2025},
  url = {https://mistral.ai/news/ministraux}
}

@article{van2025fptquant,
  title={FPTQuant: Function-Preserving Transforms for LLM Quantization},
  author={van Breugel, Boris and Bondarenko, Yelysei and Whatmough, Paul and Nagel, Markus},
  journal={arXiv preprint arXiv:2506.04985},
  year={2025}
}

@article{Widrow96,
  author={Widrow, B. and Kollar, I. and Ming-Chang Liu},
  journal={IEEE Transactions on Instrumentation and Measurement}, 
  title={Statistical theory of quantization}, 
  year={1996},
  volume={45},
  number={2},
  pages={353-361},
  doi={10.1109/19.492748}}

@article{Gersho1977,
  title={Quantization},
  author={Allen Gersho},
  journal={IEEE Communications Society Magazine},
  year={1977},
  volume={15},
  pages={16-16},
  url={https://api.semanticscholar.org/CorpusID:260498692}
}

@article{Sadegh25,
  author       = {Mohammad Sadegh Akhondzadeh and
                  Aleksandar Bojchevski and
                  Evangelos Eleftheriou and
                  Martino Dazzi},
  title        = {KurTail : Kurtosis-based {LLM} Quantization},
  journal      = {CoRR},
  volume       = {abs/2503.01483},
  year         = {2025},
  url          = {https://doi.org/10.48550/arXiv.2503.01483},
  doi          = {10.48550/ARXIV.2503.01483},
  eprinttype    = {arXiv},
  eprint       = {2503.01483},
  bibsource    = {dblp computer science bibliography, https://dblp.org}
}

@article{pusz1975,
title = {Functional calculus for sesquilinear forms and the purification map},
journal = {Reports on Mathematical Physics},
volume = {8},
number = {2},
pages = {159-170},
year = {1975},
issn = {0034-4877},
doi = {https://doi.org/10.1016/0034-4877(75)90061-0},
url = {https://www.sciencedirect.com/science/article/pii/0034487775900610},
author = {W. Pusz and S.L. Woronowicz}
}

@misc{chen2026wush,
      title={WUSH: Near-Optimal Adaptive Transforms for LLM Quantization}, 
      author={Jiale Chen and Vage Egiazarian and Roberto L. Castro and Torsten Hoefler and Dan Alistarh},
      year={2026},
      eprint={2512.00956},
      archivePrefix={arXiv},
      primaryClass={cs.LG},
      url={https://arxiv.org/abs/2512.00956}, 
}
